\documentclass[twocolumn, switch]{article}
\usepackage{PRIMEarxiv}

\usepackage[utf8]{inputenc} 
\usepackage[T1]{fontenc}    
\usepackage{url}            
\usepackage{booktabs}       
\usepackage{amsfonts}       
\usepackage{nicefrac}       
\usepackage{microtype}      
\usepackage{lipsum}
\usepackage{fancyhdr}       

\usepackage{multicol}

\usepackage{microtype}
\usepackage{amssymb}
\usepackage[colorlinks,urlcolor=blue,linkcolor=blue,citecolor=blue]{hyperref}

\usepackage{color,array, colortbl}
\usepackage{comment}
\usepackage{graphicx}
\usepackage{caption}
\usepackage{subcaption}
\usepackage{ulem}
\usepackage{multirow}

\usepackage{soul}
\usepackage{color}
\usepackage[usenames,dvipsnames]{xcolor}

\title{The path towards contact-based physical human-robot interaction}

\author{
  Mohammad Farajtabar \\
  Department of Mechanical and Manufacturing Engineering \\
  University of Calgary \\
  \texttt{mohammad.farajtabar@ucalgary.ca} \\
   \And
  Marie Charbonneau \\
  Department of Mechanical and Manufacturing Engineering \\
  University of Calgary \\
  \texttt{marie.charbonneau@ucalgary.ca} \\
}

\begin{document}
\begin{sloppypar}
\twocolumn[\begin{@twocolumnfalse}
\maketitle

\begin{abstract}
With the advancements in human-robot interaction (HRI), robots are now capable of operating in close proximity and engaging in physical interactions with humans (pHRI). Likewise, contact-based pHRI is becoming increasingly common as robots are equipped with a range of sensors to perceive human motions. Despite the presence of surveys exploring various aspects of HRI and pHRI, there is presently a gap in comprehensive studies that collect, organize and relate developments across all aspects of contact-based pHRI. It has become challenging to gain a comprehensive understanding of the current state of the field, thoroughly analyze the aspects that have been covered, and identify areas needing further attention. Hence, the present survey. While it includes key developments in pHRI, a particular focus is placed on contact-based interaction, which has numerous applications in industrial, rehabilitation and medical robotics. Across the literature, a common denominator is the importance to establish a safe, compliant and human intention-oriented interaction. This endeavour encompasses aspects of perception, planning and control, and how they work together to enhance safety and reliability. Notably, the survey highlights the application of data-driven techniques: backed by a growing body of literature demonstrating their effectiveness, approaches like reinforcement learning and learning from demonstration have become key to improving robot perception and decision-making within complex and uncertain pHRI scenarios. This survey also stresses how little attention has yet been dedicated to ethical considerations surrounding pHRI, including the development of contact-based pHRI systems that are appropriate for people and society. As the field is yet in its early stage, these observations may help guide future developments and steer research towards the responsible integration of physically interactive robots into workplaces, public spaces, and elements of private life.
\end{abstract}

\keywords{Physical human-robot interaction, Robot safety, Robot sensing systems, Robot learning, Motion planning, Compliant control, Robot Ethics}
\vspace{0.8cm}
\end{@twocolumnfalse}]


\section{Introduction} \label{introduction}
While the integration of robots into human personal and social life is not universal yet, there is a noticeable trend towards expanding their role and presence. Considering the world's aging populations and increasingly personalized lifestyles, alongside growing computational capacity, plus the emergence of artificial intelligence and machine learning (\textbf{AI/ML}), it becomes more and more probable that not only industrial, but also service and domestic robots will significantly impact our lives. 
Robotic technology is making its way into individuals' social lives in different ways, such as social robots intended for companionship or interactive robots in public settings.
Domestic robots performing household tasks, personal assistant robots, and wearable robotic devices designed to enhance daily activities are reshaping the way people approach their daily routines ~\cite{rawassizadeh2019manifestation, henschel2021makes}.

From an industrial perspective, for example in the manufacturing, logistics and warehouse industries, the introduction of robotics has altered industrial paradigms, while contributing to the creation and restructuring of numerous jobs. Robots can bring efficiency, quality, consistency and safety to production lines by reorganizing processes and lowering costs~\cite{de2008atlas, castro2021trends}. Similar impacts are soon expected in the agriculture, transportation, service, medical and retail industries, which are now experiencing a growing emergence of robotic technologies~\cite{ben2018robots}. While simple, repetitive tasks that pose safety and health risks to people may be taken over by robots, humans still have significant roles to play. 
Nonetheless, over time, human involvement will require different sets of skills and responsibilities~\cite{evjemo2020trends, 9024658}.

When robots are programmed to perform hazardous tasks, they contribute to making work generally safer for humans~\cite{de2008atlas}. With the use of robots in a common environment with humans becoming more widespread, new challenges are introduced in the escalatingly more complex field of human-robot interaction (\textbf{HRI}).
HRI can be classified as `social' (\textbf{sHRI}), for example when it consists of distanced visual or auditory and vocal interaction, it can be classified as `physical' (\textbf{pHRI}) when a robot physically interacts with humans, or it can be at once both social and physical~\cite{walther2014classification}. Whether a robot is made socially or physically closer to humans, ensuring safety becomes critical. Currently, the most prominent ways to address physical safety directly involve robot perception, planning and control, while psychological safety has yet to be addressed through ethical and psychosocial frameworks. 
Each of these fields (perception, planning, control, ethics) are also central to the development of robots that can effectively perform their intended task while harmoniously physically interacting with humans.

pHRI, and in particular \textbf{\textit{contact-based} pHRI} involving active, deliberate contact between humans and robots, is still in the early stages of development. There is however a growing, scattered body of literature on the topic.
The contribution of this paper is therefore to provide a comprehensive and interconnected perspective on pHRI, encompassing technical, safety, and ethical challenges and considerations. The paper aims to provide a holistic understanding of the current research in contact-based pHRI, including background knowledge for new researchers in the field, while calling attention to challenges that have yet to be tackled. 

Numerous aspects come together to make robots that can directly physically interact with humans. This review will kick off by defining pHRI within the context of HRI in Section~\ref{HRI_definitions}. Section~\ref{safety} will then delve into the topic of safety, which we consider as the foremost challenge in pHRI, to explore how it has so far been addressed through strategies involving perception, planning and control, and leveraging the power of AI/ML. The paper then surveys the different ways in which robots that can effectively physically interact with humans have been developed.
In Section~\ref{perception}, an extensive discussion of perception, including sensor development and human intent detection is presented. Planning approaches that have been proposed to carry out pHRI applications are explored in Section~\ref{planning}, and Section~\ref{control} covers robot motion control for contact-based pHRI, with special attention to the realization of variable compliance control schemes, stability, and AI/ML approaches. Section~\ref{computational_enhancement} discusses computational considerations, covering aspects related to both software and hardware perspectives. We additionally examine the ethical considerations that must be taken into account when designing and deploying robots that closely interact with humans in Section~\ref{ethics}. Finally, Section~\ref{conclusion} summarizes the current state of research in contact-based pHRI, leading to our identifying future trends and challenges that have yet to be addressed.


\section{Definitions of human-robot interaction} \label{HRI_definitions}
\begin{figure}[htbp]
\centering
    \begin{subfigure}[b]{0.9\linewidth} 
         \centering
         \includegraphics[width=\textwidth]{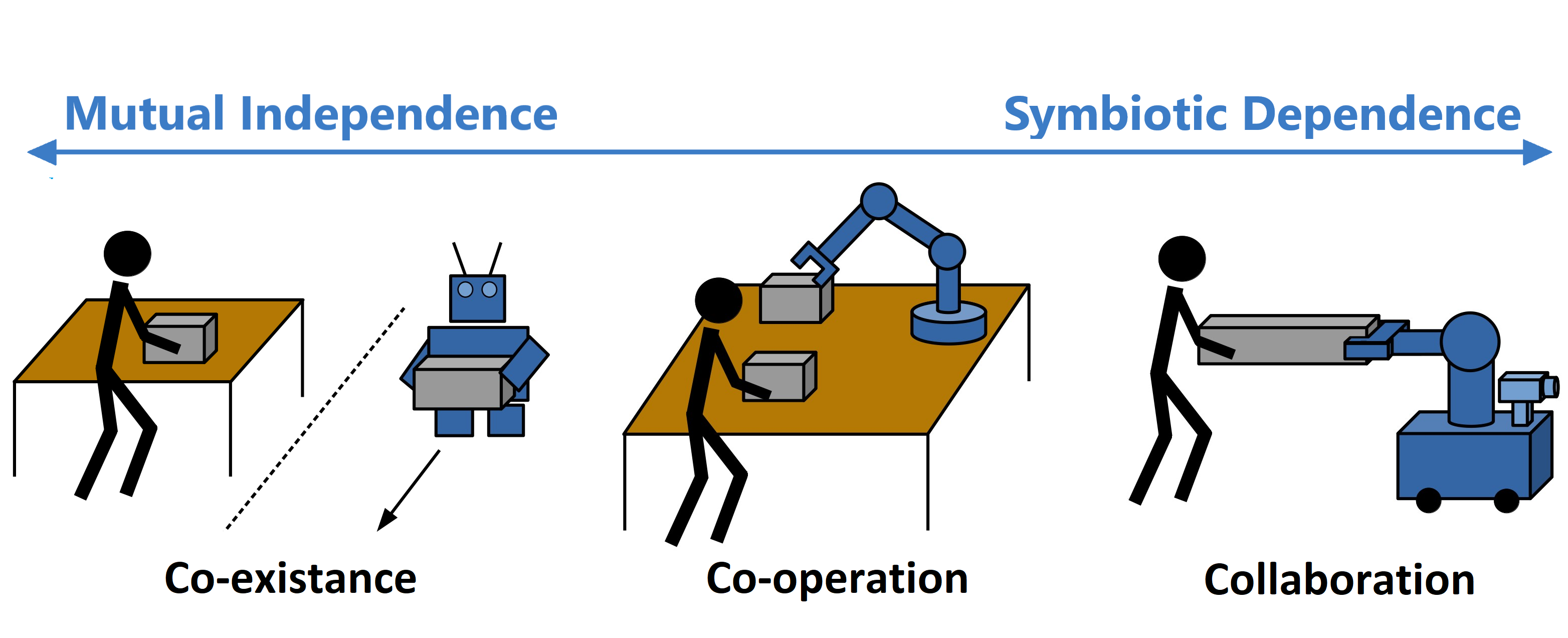}
         \caption{Classification by the degree of independence}
         \label{fig:HRI_degree_of_independence}
     \end{subfigure}
     \hfill
     \begin{subfigure}[b]{0.7\linewidth} 
         \centering
         \includegraphics[width=\textwidth]{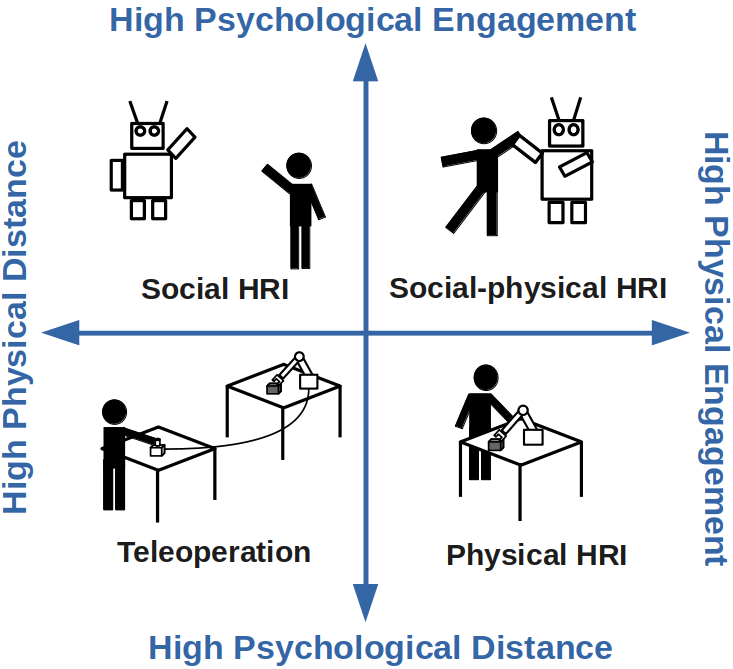}
         \caption{Classification by the degree of engagement}
         \label{fig:HRI_degree_of_engagement}
     \end{subfigure}
\caption{Classifying HRI based on two factors. a) the degree of independence between human and robot actions, including co-existence: humans and robots occupy separate workspaces without any interference, co-operation: humans and robots share a workspace while working on their respective tasks, and collaboration: humans and robots share a workspace and simultaneously work together on a common task. b) the degree of physical and psychological engagement in the interaction. The four quadrants mark the division between social, physical, social-physical HRI and teleoperation. Each quadrant encompasses various forms of interaction along the co-existence-co-operation-collaboration spectrum.}
\label{fig:HRI_classification}
\end{figure}

HRI can be defined as the study of how people interact with robots. It includes the design, development, and evaluation of systems that allow people to interact with robots in a natural and intuitive way. HRI research covers a wide range of topics, including human-robot communication, user interface design, robot perception, robot autonomy, safety and ethics. The goal of HRI is to develop robots that can work alongside humans in a variety of settings, such as homes, workplaces, and public spaces. In general, HRI can be categorized from two perspectives: (A) the degree of independence between human and robot actions, or (B) the degree of engagement between a human and a robot. 
The next two subsections detail these two classification themes, and Fig.~\ref{fig:HRI_degree_of_independence} provides a graphical representation.

\subsection{HRI categorized by degree of independence}
As shown in Fig.~\ref{fig:HRI_degree_of_independence}, HRI can be divided among three broad categories of interaction: co-existence, collaboration, and co-operation, as explicitly introduced in~\cite{muller2017subjective}, and further used in~\cite{de2008atlas, 8907351, tsarouchi2016human, zacharaki2020safety}
\textbf{Co-existence} refers to humans and robots operating in the same environment without mutual interference, emphasizing individual independence. An example of co-existence might be robots transporting goods in a warehouse where humans are doing the packing.
\textbf{Co-operation} 
refers to coordinated teamwork with clearly defined roles and responsibilities for a human and a robot working within a shared workspace, but who do not engage in simultaneous work on the same item. An instance of this can be observed in a production line scenario where both partners sequentially manipulate an object.
\textbf{Collaboration} involves humans and robots actively working together: sharing information, resources and decision making processes to achieve a shared goal, thus entailing a symbiotic dependence. An example of this could be a robot assisting a human by handing over objects or jointly carrying goods.

\subsection{HRI categorized by degree of engagement}

HRI can be categorized into remote interaction and proximate interaction, as suggested in~\cite{goodrich2008human}:
\begin{itemize}
\item Remote interaction: human and robot are separated spatially or temporally
\item Proximate interaction: human and robot are in the same location or environment
\end{itemize}
As distance can be interpreted in either physical or psychological terms, Fig.~\ref{fig:HRI_degree_of_engagement} illustrates a division of HRI into four general quadrants relating to the degree to which interactions involve psychological and physical connections.

\textbf{Social HRI (sHRI)} refers to the ability of a robot to engage with humans socially, which includes engaging in conversations and responding to emotional cues. There currently are two prevalent forms of social interaction: speech-based interaction, which involves using speech recognition and natural language processing for vocal communication, and visual-based interaction, which encompasses the recognition of gestures, body language, and the use of lights or displays for communication~\cite{yan2014survey, yang2018social, butepage2017human, akalin2021reinforcement}. For example, socially skilled robots can offer companionship to healthcare patients. A study conducted in~\cite{costa2015using} used a humanoid robot to assist children with autism in improving their body awareness and social interactions, demonstrating the potential effectiveness of robots as a tool to educate children with autism.
In entertainment venues like theme parks and museums, robots have also been employed to engage visitors and enrich their overall entertainment experience~\cite{1521743, pollmann2023entertainment}. 

\textbf{Physical HRI (pHRI)}, in the opposite corner, involves robots and humans interacting with each other\linebreak through different types of contact~\cite{de2008atlas, 8907351, tsarouchi2016human, 6840111}, within a large range of applications and industrial settings~\cite{walther2014classification}. pHRI is central to assistive technology, where it can be used to assist individuals with disabilities or older adults in achieving mobility or in carrying out daily tasks~\cite{lasota2017survey, rahimi2018neural, han2019admittance}. In healthcare, robots can play a crucial role through pHRI in medical procedures, surgeries, and physical rehabilitation~\cite{marban2019recurrent, 9594838, kim2017impedance, 8333285}. In manufacturing and industrial settings, pHRI can be leveraged to enhance productivity and safety by having robots perform dull, repetitive and dangerous tasks when working alongside humans~\cite{mohammadi2020mixed, app11093907, yao2018sensorless, 10144527}.

\begin{figure}
     \centering
    \begin{subfigure}[b]{0.35\linewidth} 
         \centering
         \includegraphics[width=\textwidth]{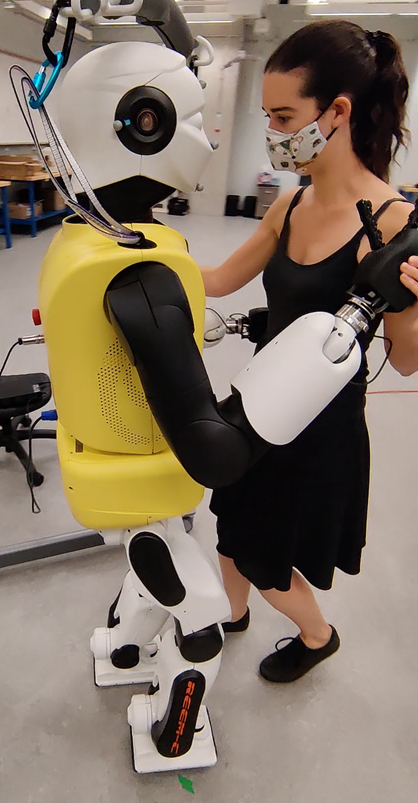}
         \caption{Direct interaction}
         \label{fig:direct}
     \end{subfigure}
     \hfill
     \begin{subfigure}[b]{0.53\linewidth} 
         \centering
         \includegraphics[width=\textwidth]{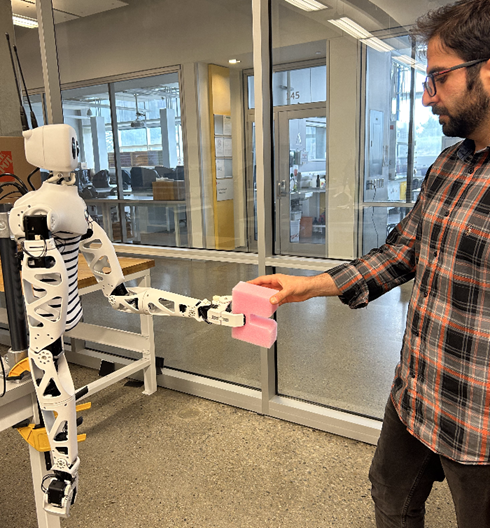}
         \caption{Handover}
         \label{fig:handover}
     \end{subfigure}
     \hfill
    \hfill
     \begin{subfigure}[b]{0.6\linewidth} 
         \centering                           
         \includegraphics[width=\textwidth]{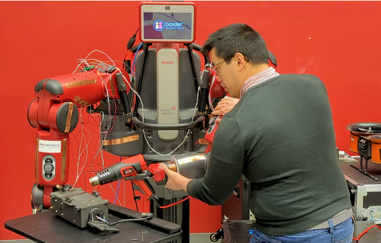}
         \caption{Co-manipulation}
         \label{fig:comanipulation}
     \end{subfigure}
     \hfill
        \caption{Three different scenarios in pHRI: a) Dancing, leveraging direct physical contact, b) Handover task, where the robot transfers an object to the human,
        c) Co-manipulation, where the human and robot collaborate to accomplish a task through direct contact. Picture reproduced from~\cite{10144527} with permission.
        }
        \label{fig:forms_of_pHRI}
\end{figure}

Within pHRI, distinct types of interaction can be identified, as shown in Fig.\hspace{0.05cm}\ref{fig:forms_of_pHRI}. 
These include: 
\begin{itemize}

\item Direct (contact-based) physical interaction: involves direct contact between human and robot, such as through touch or grasping, for example shaking hands~\cite{8794065}, dancing~\cite{7858650}, kinesthetic teaching~\cite{akgun2012trajectories}, assisting humans in industrial tasks~\cite{10144527}, in rehabilitation~\cite{han2019admittance, topini2022variable}, or in surgery\cite{kim2017impedance}.

\item Indirect physical interaction: involves humans and robots interacting through the intermediary of objects, for instance when collaborating to achieve a common goal such as manipulating or moving objects together~\cite{7759417, khoramshahi2019dynamical}, assembling a product~\cite{7989334}, or handing objects over~\cite{7139393}.

\item Proximity interaction: involves the robot and the human working in close proximity to each other, but not necessarily in direct or indirect physical contact. For example, accomplishing complementary roles to collaboratively accomplish a task~\cite{tsiakas2017interactive}, or human-aware robot navigation in common environments such as warehouse and manufacturing~\cite{9340865, 9806047}.
\end{itemize}

\textbf{Social-physical HRI} (\textbf{spHRI}) involves the integration of social and physical cues to establish human-like, engaging interactions between humans and robots. This form of interaction goes beyond verbal or visual communication by incorporating physical gestures, touch, and movement.

In  the service and hospitality industries, spHRI can lead to robots interacting and engaging with customers, taking and delivering orders and enhancing the overall customer experience~\cite{zacharaki2020safety, 1521743, pollmann2023entertainment, rozo2016learning, 9075439}.
In education, spHRI can facilitate teaching and learning of various topics.
For instance, in~\cite{7858650}, a robot is programmed to teach dancing skills, providing learners with social and physical feedback. Experiments conducted in~\cite{blancas2015effects} and~\cite{henkemans2013using} explored the use of robots in educational settings for children. 
The findings suggest that the presence of a robot acting as a teacher or tutor assistant, and which interacts socially or physically with children, can positively affect their interest in learning.

Integrating the social and physical aspects of HRI can lead to increased safety and efficiency of robots that operate alongside humans, as suggested in~\cite{8625030}. This paper introduces a spHRI framework, by combining visual perception of humans, a robot controller that safely reacts before and after contact with humans, and data from people and objects perceived in the environment.

\textbf{Teleoperation} consists of a more distanced type of HRI, where humans interact with robots remotely, e.g., remotely performing inspections
through a telepresence robot, as described for example in~\cite{de2008atlas, 8907351, 6840111}. Teleoperation can be merged into pHRI when augmented with haptics, or into sHRI when used to communicate with people through the intermediary of the teleoperated robot.

One of the main challenges currently limiting the development of applications involving physical HRI is safety. When humans and robots engage in any of the interaction types described above, ensuring human safety indeed becomes a major challenge and a significant aspect of HRI~\cite{zacharaki2020safety}. The next section delves deeper into that topic.


\section{Safety} \label{safety}

Prevention of accidents and injuries, whether physical or psychological, has long been and still is a key challenge in HRI. 
Traditionally, ensuring physical safety has required preventing any unintentional or unwanted physical contact between humans and robots, for example by establishing safety zones that isolate robots from humans. 
When physical contact is unavoidable or required for a specific task, as in some pHRI scenarios, the typical approach is to keep forces applied by the robot within a safe range. By ensuring psychological safety, one aims to create stress-free and comfortable HRI, for instance through the robot's appearance, embodiment, gaze, speech, posture, and adherence to social conventions~\cite{de2008atlas, zacharaki2020safety, lasota2017survey, robla2017working}.

Recent advancements in human physical and psychological factors, materials, sensing technology, motion planning and control, as well as the integration of AI/ML, have the potential to expand robot capabilities toward physically safe HRI
~\cite{lima2019artificial, semeraro2023human}. 

Researchers so far have formalized levels of damages and accidents that can occur through HRI~\cite{robla2017working, vasic2013safety}, along with suggested strategies for accident prevention and safety enhancement. After surveying literature on pHRI, one may categorize the array of proposed approaches into four main types of safety strategies, each addressed in the next four subsections: (\ref{sub:safety_design}) robot design, (\ref{sub:safety_prediction}) human prediction, (\ref{sub:safety_planning}) motion planning, and (\ref{sub:safety_control}) control.

\subsection{Robot design} \label{sub:safety_design}

Robot design 
encompasses the physical design of a robot, its components, and the environment in which it operates~\cite{papetti2022human, boschetti2022human, gualtieri2022development}. Designing for safety can then be addressed from multiple angles, as detailed in the following subsubsections. One can enhance safety from the perspective of (\ref{subsub:ergonomics}) ergonomics, as well as (\ref{subsub:social_factors}) social and psychological factors. In addition, one can minimize the risk of injury caused by collisions by designing (\ref{subsub:material}) softer, lighter, and compliant robots, or minimize the risk of collisions through (\ref{subsub:sensors}) sensing systems.

\subsubsection{Ergonomics and physical human factors} \label{subsub:ergonomics}

When it comes to robotics, ergonomics is typically focused on optimizing human well-being and robot performance through the analysis of interactions between humans and robots, often targeting to minimize the risk of work-related musculoskeletal disorders. To do so, an ergonomic assessment of the robot-human system throughout the design process will be crucial. To that effect,~\cite{maurice2017human} introduced a method for detailed ergonomic assessments of collaborative robots, identified influential parameters to improve ergonomics and defined a robot design optimization algorithm based on their analysis. 
A different methodology optimizing robot hardware parameters with the objective of ergonomically minimizing energy expenditure, was developed and applied to a collaborative payload lifting task in~\cite{10000222}. 

To design ergonomic social and service robots, and make the environment in which they operate also ergonomic, a cross-disciplinary approach, bringing together roboticists, architects, and designers, was taken in~\cite{sosa2018robot}, integrating a variety of factors such as human size, aesthetics, appropriateness and simplicity. 
Physical ergonomics is crucial for improving pHRI in repetitive tasks, such as industrial assembly.
Guidelines are proposed in~\cite{gualtieri2020safety} for the design of safe and efficient human-centered collaborative assembly workstations, based on international standards, research, and real-world cases. In particular, the guidelines seek to mitigate
1) upper-body load during repetitive tasks,
2) whole-body load when lifting or lowering objects,and 3) whole-body strain when maintaining working postures.
As pHRI is developed further, ergonomic close physical interactions will remain essential for physical safety, just as much as psychological safety will need attention. 

\subsubsection{Social and psychological human factors} \label{subsub:social_factors}

Social and psychological factors may often be overlooked in robot design. However, they play crucial roles in ensuring \textbf{perceived safety},
encompassing the feeling of safety and security conveyed to humans during pHRI~\cite{rubagotti2022perceived}. It arises from factors such as comfort, predictability, transparency, sense of control, and trust~\cite{akalin2022you}. Approaches aimed at enhancing psychological safety in pHRI typically focus on fulfilling these factors~\cite{lasota2017survey, 4581481}.
Findings from~\cite{lim2021social, alzahrani2022exploring} suggest that individuals from diverse cultures may place varying levels of trust in robots. According to these studies a multitude of cultural aspects, including communication, attitudes, values, beliefs, expectations, technology levels
must be considered to enhance trust and psychological safety in pHRI. 
The review in~\cite{lu2022mental} investigated workers' mental stress and safety awareness in human-robot collaboration, finding that it is affected by robot size, speed, trajectories and contacts with the robot.
The impact of different robot behaviours on human discomfort, perceived safety, sense of control and distrust was explored in~\cite{akalin2022you}, with results indicating that perceived safety is also influenced by individual human characteristics (e.g., personality and gender), and that physiological signal data can be effective in measuring perceived safety.

Thus, robot design, in terms of physical characteristics and interactive behaviours, significantly impacts perceived safety. One way to effectively increase how intrinsically safe a robot is to be around, is through robot structural design.

\subsubsection{Structural design and material selection} \label{subsub:material}

Enhancing safety involves prioritizing the development of robot hardware that is both user-friendly and inherently safer. In instances of collision, employing a compliant robot structure proves beneficial in minimizing potential harm to humans. Within the realm of robot hardware production, three key design elements are commonly emphasized to mitigate collision energy, whether interactions are deliberate or accidental: mechanical compliance in the robot's links and actuators (referred to as passive compliance), the utilization of soft and energy-absorbing robot skin, and the adoption of lightweight manufacturing techniques for robots~\cite{pervez2008safe}.

\textit{Mechanical compliance in robot links and actuators}\\
Safety in pHRI can be enhanced with the introduction of tunable stiffness robot links, as proposed in~\cite{she2016design}, where servo motors are used to adjust the stiffness of actuated four-bar linkages.
Impact tests on this approach have revealed a significant reduction in acceleration and head injury criteria, indicating improved safety for operators during collisions.

A compliant actuator can be defined as an actuator designed with a mechanically low impedance (for example through a spring), which therefore permits deviations from its equilibrium position, with 
minimal force or torque in response to external forces. In contrast, a stiff actuator would be designed with a high mechanical impedance, allowing it to remain fixed once it reaches a specific position, regardless of external forces (within the limits of the forces and torques it can widthstand)~\cite{van2009compliant}. A notable example of compliant actuation is the use of series elastic actuators (\textbf{SEAs}), which has been proposed in collaborative robots to absorb collision energy and make interactions safer. Broadly speaking, SEAs are composed of a spring connected in series with a stiff actuator. The compliance of SEAs is fixed and determined by the spring constant, which cannot be adjusted during operation. In particular, this setup facilitates force control~\cite{zinn2004playing, pratt1995series}. 
Another example of compliant actuators is the variable stiffness actuator (\textbf{VSA}), designed with adjustable stiffness, rendering them suitable for safe pHRI. Their energy-efficient nature and adaptability to various tasks, environments, or conditions make them a promising option for robot actuation. This type of actuators generally comprise three pulleys, with two of them independently controlled by motors, linked to the joint via a timing belt~\cite{1570172, bicchi2005variable}.

Passive compliance can also be attained through backdrivable transmissions, where external forces exerted between human and the robot is reflected in motor currents. This setup facilitates robust torque control, since the motor functions as a torque sensor. By colocating the sensor and actuator, it notably alleviates dynamic stability issues encountered in force control. Direct-drive (\textbf{DD}) and quasi direct-drive (\textbf{QDD}) actuators are two examples of backdrivable transmissions proposed in~\cite{8794236, 7403902}.

\textit{Energy-absorbing robot skin}\\
Various implementations of energy-absorbing robot skins have been introduced over time, such as employing viscoelastic coverings~\cite{robla2017working, 525724, lim1999collision}, to reduce the potential impacts of collisions, while at the same time offering a tactile experience reminiscent of human skin. In~\cite{7353705}, authors devised a soft skin incorporating pressurized airbags connected to a pressure sensor to identify the force exerted on the robot arm covered by the skin.
Another instance of a soft skin equipped with tactile sensors to detect touch is introduced in~\cite{chang2015interaction}. The robot's soft, hypoallergenic fur-covered skin ensures the safety of pHRI.
Utilizing inherently soft skins without soft sensing components offers an alternative for swift and safe interaction, as there is no delay associated with force sensing. A soft inflatable robotic arm, such as the one introduced in~\cite{6907362}, may be made inherently safe without requiring external force sensors, resulting in reduced delay in the control system.

\textit{Lightweight robot manufacturing techniques}\\
Lastly, the design of lightweight structures, such as introduced in~\cite{bicchi2004fast, 5756872}, is another safety-oriented consideration aimed at minimizing injuries in the event of collisions with humans.
Decreasing robot mass results in decreased momentum, thus reducing impact forces when accidental contact or collision occurs in pHRI~\cite{de2006collision}. As a result, using lightweight robots contributes to safety in pHRI by presenting a lower risk of injury in comparison to heavier, more traditional, robots.

From there, to evaluate the extent to which a robotic system is safe, psychologically and physically, and to generate safe robot behaviours, sensors will be the crucial pieces of equipment to consider.

\subsubsection{Sensors} \label{subsub:sensors}

Sensors are critical in ensuring safe pHRI. A range of human physiological sensors may be used to evaluate perceived safety during pHRI~\cite{rubagotti2022perceived}. Robot sensors, for their part, enable environment awareness and detection of human presence. 
Tactile, pressure, 6-axis force/torque (\textbf{F/T}) and joint torque sensors, as well as cameras, are commonly employed in this sense.
Tactile sensors allow perception of human touch, while pressure sensors measure a force applied over a given area. An F/T sensor measures the resulting 3D forces and torques from wrenches applied to robot body parts situated distally from the sensor, while a joint torque sensor measures the 1D torque applied at a robot joint as a result of wrenches applied distally. Cameras, systems composed of stereo and/or range cameras as in~\cite{rybski2012sensor}, or RGB-D sensors, can be used to capture 3D information about a robot's surroundings. These systems, when installed on a robot or in the environment, capture 2D images alongside depth information, allowing to map the distance between the sensor and objects in the environment, ultimately aiding in estimating human pose. 

Artifical skin sensors can also be developed to measure interaction forces on robot body regions, such as the pressure-sensitive skin introduced in~\cite{fritzsche2011tactile}. Integrating a pressure sensitive layer with an energy absorbing layer, it can conform to complex shapes and reduce the risk of injury while measuring contact pressure during pHRI.
In~\cite{8793258}, an artificial robot skin is introduced, comprising an array of tactile sensor cells capable of detecting 3D acceleration, force, proximity, and temperature. 
Alternatively, \cite{maiolino2013.iCubSkin} introduced flexible tactile sensors inspired by techniques from the clothing industry.

The information provided by sensors can therefore be used for the implementation of safety measures that prevent potential harm to humans~\cite{10144527, vasic2013safety, 9197365, 7346450}, for instance through prediction of human movements, robot motion planning and control, as covered in the next subsections.

\subsection{Human detection and motion prediction} \label{sub:safety_prediction}

To ensure safety during pHRI, planning and control strategies often rely on an explicit evaluation of potential danger to humans, for example based on factors that influence collision impact forces such as relative distance and velocity between the robot and human and robot inertia~\cite{haddadin2012making}. From there, the ability to perceive human presence and behaviours around the robot can be another essential aspect to improving safety.

Computer vision and learning approaches have been used to build a human model that estimates human pose and intention~\cite{morato2013safe, rybski2012sensor}. In~\cite{kulic2007pre}, a methodology is proposed to integrate vision and physiological sensor-based data on the user's position and physiological responses into medium and short-term safety strategies. Another framework to generate safe robot motions is introduced in~\cite{mainprice2013human}, based on early human motion recognition using a Gaussian Mixture Model (\textbf{GMM}), and human motion prediction using Gaussian Mixture Regression (\textbf{GMR}). 
In~\cite{5980248}, the authors proposed a method for predicting human arm motion for safe interaction using a red marker attached to the arm of a human tracked by cameras. Using data from experiments, a topological map is created, and arm velocities are estimated through a hidden Markov model (\textbf{HMM}). 
Likewise,~\cite{9300047} introduces a framework for predicting human arm motion during a reaching task. It integrates partial trajectory classification and human motion regression, enabling action recognition and trajectory prediction before completion of the movement. 
Combining computer vision with deep learning can also help handle the complex nature of human models.
~\cite{CHOI2022102258} presents a mixed reality system for safe human-robot collaboration using deep learning and digital twin technology. It measures real-time safe distances, provides task assistance, and integrates mixture regression with safety monitoring using RGB-D cameras.

In particular, detecting human intention and predicting motion through gaze plays a crucial role in fostering successful and safer interactions. Gaze offers nuanced cues that facilitate smoother communication between individuals. By integrating predictions based on human gaze, it becomes possible to gauge the level of engagement during human-robot interactions, enabling the robot to anticipate the intentions or objectives of the human participant~\cite{admoni2017social, 8593580}. In collaborative interactions between humans and robots, the utilization of gaze tracking and eye monitoring can contribute to anticipating the human operator's workload and performance levels throughout the task~\cite{upasani2023eye, haji2018exploiting}.

Along the same line,~\cite{10144527} integrates vision and tactile sensors information and identifies touch location, human pose, and gaze direction, and uses that information to train a machine learning algorithm which classifies intentional and unintentional touch. Alternatively, in~\cite{8793657}, signals from F/T sensors are combined with motor currents to differentiate accidental collisions from intentional ones. This information has been shown in both~\cite{10144527,8793657} to be instrumental in appropriately adapting robot behaviour for safe pHRI, for instance through motion planning and control.

Collision-based detection, i.e., identifying when a robot comes into contact with humans in its workspace, can also be crucial for safety. For instance, the authors in~\cite{4650764}, developed and evaluated two distinct collision detection approaches: a disturbance observer, and a robot torque observer. Broadly speaking, collision detection methods based on observers, like those mentioned above or a momentum observer, offer utility as they remove the need to solve the inverse dynamics problem~\cite{li2021nonlinear}. 
Once a collision has been detected, the actions of the robot must then be adjusted for safety, making use of appropriate motion planning and control approaches. From there, human state information is vital for the introduction of a safety strategy, typically relying on planning.

\subsection{Motion planning} \label{sub:safety_planning}

Going beyond collision prevention at the control level, real-time human-aware motion planning has been shown to improve physical and psychological safety in HRI~\cite{lasota2017survey, 6899348, lasota2015analyzing}.
For instance,~\cite{6197743} introduces a human-aware planner for robot handover tasks, which determines handover location based on factors including human kinematics, field of vision and personal preferences, thus reducing human cognitive load and increasing comfort and efficiency. 
To enhance safety and robot predictability, a motion planning cost function is proposed in~\cite{7487584}, taking into account avoidance of previously occupied workspace by human, and robot motion consistency.
Instead, a human model is integrated into robot path planning in~\cite{9913366}, with the objective of minimizing path execution time while slowing down and stopping when humans are in proximity. This is achieved by defining speed limits as configuration-space cost functions based on observed and predicted human states.

Although most planning approaches focus primarily on pre-collision strategies, such as in~\cite{de2007reactive, 8004480}, when contact is expected or unavoidable, motion planning must adeptly manage contacts to ensure safety. Traditionally, this has been viewed primarily as a control problem, but integrating contact handling into trajectory planning can lead to more sophisticated and human-friendly interactions~\cite{haddadin2013almost}. The idea introduced in~\cite{4650764} for post-collision robot reaction involves retaining the original motion path, while also incorporating compliant behaviour through adjustments in the timing of the intended trajectory, in response to a collision. 

In~\cite{haddadin2011dynamic}, a reactive planner is proposed, in addition to pre-collision planning. This reactive planner determines the course of action during a collision: end-effector motion is stopped and, a new internal model is created based on the time evolution of contact forces, and then the new destination for retraction is obtained.

A model-based trajectory planning algorithm is introduced in~\cite{9488306} for rehabilitation and physical therapy robotics, alongside a framework incorporating a human musculoskeletal model to evaluate patient conditions and perform physical therapy movements safely with a wide range of robot motions. 
An example of safe real-time motion planning allowing legged humanoid robots to safely engage in physical interaction with humans is furthermore introduced in~\cite{9034996}, focusing on the dynamical planning of bipedal walking trajectories and adapting steps for successful push recovery.

Nonetheless, safe and human-aware robot motion planning may not be sufficient to guarantee safe pHRI: motion control also needs to be considered.

\subsection{Control} \label{sub:safety_control}

Several control strategies can be adopted for safety during pHRI, including acting~(\ref{subsub:precollision}) before and~(\ref{subsub:postcollision}) after a collision occurs, or~(\ref{subsub:compliance}) ensuring robot compliance at all times, and~(\ref{subsub:stabilization}) stabilization of unstable robots.

\subsubsection{Pre-collision strategies} \label{subsub:precollision}

These techniques involve monitoring either the human, the robot, or both, and adjusting robot control parameters before a collision or contact occurs. The objective is to proactively modify the robot's behaviour to prevent potential accidents and ensure a safe HRI environment~\cite{zacharaki2020safety}. 
For example, \cite{kulic2007pre} designed a motion planning and control approach which relies on an explicit estimate of the danger level during the interaction, taking into account measured human heart rate, skin conductance and contraction of facial muscles.

Adopting a different approach, \cite{haddadin2012making} proposes a controller which maintains robot velocity within dynamic bounds, determined based on robot dynamics and configuration, as well as an estimate of injuries that may be incurred in the context.
Seeking instead to balance safety and productivity while leveraging robot redundancy,~\cite{7079531} proposes a controller in which robot velocity is scaled within a safety region defined based on its current velocity and braking distance. 

Extensive efforts to prevent collisions and minimize the associated risks may not entirely eliminate the occurrence of collisions. Consequently, post-collision strategies are just as essential to reduce the severity of injuries.

\subsubsection{Post-collision strategies} \label{subsub:postcollision}

Post-collision safety strategies may include designing robots with compliant and soft materials that can absorb impact forces and reduce the risk of injury upon collision with a human, as discussed in~\ref{subsub:material}. Another approach may be to design robots with sensor systems that can detect collisions or contacts in real-time. Upon detection, the robot can stop moving or dynamically adjust its behavior to minimize any harm or damage~\cite{robla2017working}. 

A common approach to mitigate collisions is to adjust control parameters based on joint torque and encoder signals, as presented in~\cite{oleinikov2021safety, vick2013safe, bian2018improving}. One can also leverage signals from an F/T sensor mounted at the robot end-effector: for instance, \cite{8793657} uses measured F/T signals in combination with motor currents, to differentiate accidental collisions from intentional contacts. Robot response is then adapted in consequence to ensure safety, e.g., by interrupting any motion after accidental collision, or by ensuring robot compliance to intentional contact as described next.

\subsubsection{Compliant control} \label{subsub:compliance}

In articulated collaborative robots, signals from F/T sensors, joint encoders and torque sensors as mentioned above, or even artificial skins~\cite{silvera2015artificial, li2022multifunctional}, may be instrumental in developing robot motion controllers that are compliant to physical interactions and safe for pHRI (i.e., active compliance).

Passivity, or maintaining the energy stored within system elements (comprising kinetic and potential energy) lower than the input energy, is crucial for ensuring stability, and consequently, safety~\cite{6631284}.
Within this context, passivity-based approaches designed to guarantee compliant controller stability during physical interactions are highly useful for safety. However, passivity alone may not directly address the need to maintain safe robot configurations, velocities, power and interaction forces. This is for example addressed in~\cite{cortez2021safe}, where a robot controller is designed to maintain motion within safe position and velocity regions, while ensuring passivity during pHRI. Alternatively, in~\cite{chen2022human} and~\cite{8685113}, a variable admittance control approach is introduced based on principles of passivity, allowing to adjust admittance parameters (damping, inertia, and stiffness) in such a way as to limit interaction forces, thus enhancing safety during pHRI.

Compliant control is currently the most common control approach in pHRI, being employed beyond the single purpose of ensuring safety. For this reason, Section~\ref{control} will delve deeper into the subject.

\subsubsection{Stabilization} \label{subsub:stabilization}

While the strategies outlined in subsections~\ref{subsub:material}, ~\ref{subsub:sensors}, ~\ref{sub:safety_prediction}, ~\ref{sub:safety_planning}, and ~\ref{sub:safety_control} of the current section on Safety contribute to the safety of both the human and robot, another strategy consists in primarily enhancing the safety of the robot itself.
Such a strategy is especially relevant for locomoting robots that are not passively stable, such as humanoids and ballbots.
Introducing compliance (either passive or controlled) in such robots is likely to compromise their balance, along with their ability to prevent falls or tipping over, especially when subjected to external forces.
Therefore, incorporating stabilizing control algorithms (e.g., stabilizers such as introduced in~\cite{Zhou2014Stabilizer}) into the control framework is essential to ensure safety during pHRI, preventing the robot from collapsing and either breaking itself or injuring a human through its fall.
Exploring this issue,~\cite{tirupachuri2020towards} presented a control strategy, adapted for a human-assisted robot sit-to-stand scenario, which takes into account interaction forces applied to a compliantly torque-controlled humanoid robot, when computing control inputs that track a desired center of mass trajectory.
Contrastingly,~\cite{7139727} introduced an impedance control framework allowing a ballbot to maintain balance in a robot-assisted human sit-to-stand scenario. Another framework was proposed in~\cite{8968546} for indoor service ballbots, offering physical aid and dynamic guidance to individuals navigating congested and tight spaces. To maintain balance, the robot employs a balancing PID controller to mitigate disturbances and uphold the desired roll and pitch angles of the body.

In~\cite{kobayashi2022whole}, an approach to close multi-contact physical interaction between human and humanoid robots was introduced. The proposed controller, based on the divergent component of motion~\cite{englsberger2015DCM}, maintains balance while stepping by considering pHRI forces as disturbances.

As control makes heavy use of perception data, the topic of perception will first be covered in Section~\ref{perception}.


\section{Perception} \label{perception}
In HRI, awareness can be defined as the human's perception and understanding of a robot’s location, activities, status, and surroundings and vice versa~\cite{1243931}. The robot relies on information about the human's commands or instructions to guide its actions within specific conditions and limitations. Insufficient awareness significantly hampers the level of interaction, resulting in a notable decrease in overall task performance, when human-robot collaboration is required~\cite{1243931}. To organize our analysis, we divide the robot perception and control chain in pHRI into four main modules~\cite{russell2016artificial, durrant2012integration}:
\begin{enumerate}
\item[A.] Sensor development: gather information from the environment through various sensors
\item[B.] Sensory data integration: build a representation of the environment, users, and interaction forces
\item[C.] System modelling: develop models to analyze the integrated data and predict future states of the system
\item[D.] Planning and control: implement robot decision policies based on optimization, and command robot motion to achieve a given task
\end{enumerate}

The following subsections address modules A, B, and C to provide a comprehensive exploration of perception in pHRI, while module D will be the topic of Sections~\ref{planning} and~\ref{control}.

\subsection{Sensors for pHRI}

This subsection covers the development and use of the main sensors used for contact-based pHRI. A handful of sensors count as the most commonly used to measure physical interactions: joint torque, F/T, tactile, cameras and depth sensors, but other types of sensors have also shown to be useful. 

Figure~\ref{fig:tactile, F/T sensors, RGB_D} shows sensor configurations for pHRI on different robots.

Data from \textbf{joint torque sensors} relates directly to the forces exchanged in pHRI, e.g., as leveraged in~\cite{marban2019recurrent}.
When a robot is equipped with joint torque sensors, it is common practice to use the measured torques as feedback in compliant joint torque control laws adapted for pHRI, as surveyed in~\cite{Haddadin2016.pHRI}, and implemented in~\cite{Grunwald2003.TorqueControl,topini2022variable}. 

\textbf{F/T sensors} mounted on robot end-effectors are widely employed to measure contacts during pHRI and to implement compliant control algorithms~\cite{5152664, 8328912, 7989338, 8307450}. 

\begin{figure}
     \centering
    \begin{subfigure}[b]{0.3\linewidth} 
         \centering
         \includegraphics[width=\textwidth]{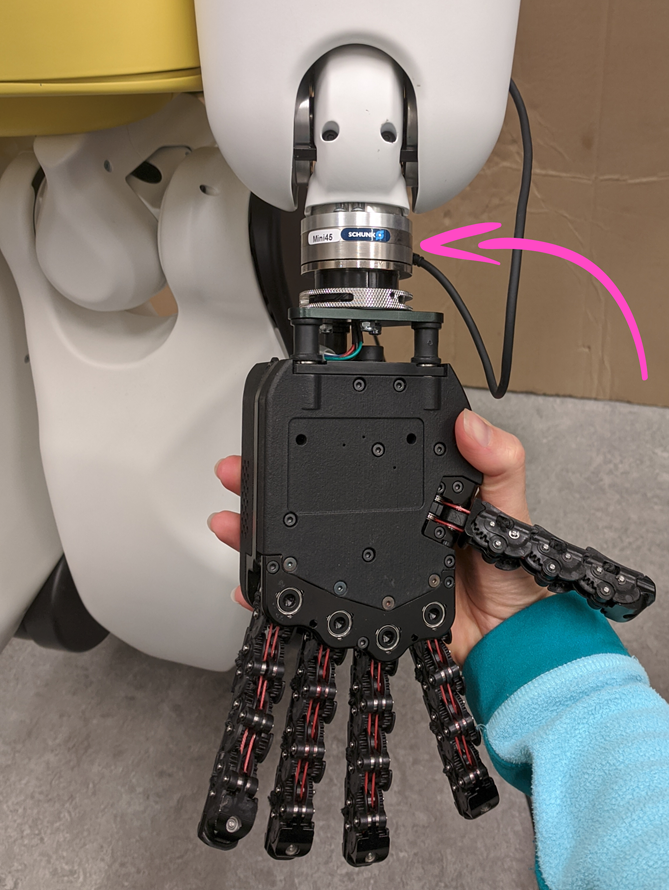}
         \caption{F/T sensor}
         \label{fig:F/T_sensor_module}
     \end{subfigure}
     \hfill
     \begin{subfigure}[b]{0.3\linewidth} 
         \centering
         \includegraphics[width=\textwidth]{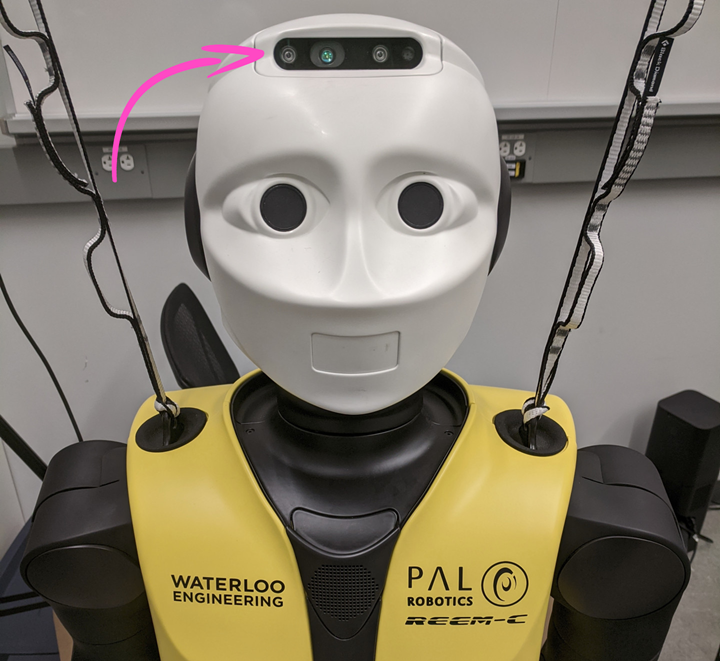}
         \caption{RGB-D sensor}
         \label{fig:RealSense}
     \end{subfigure}
     \hfill
    \hfill
     \begin{subfigure}[b]{0.3\linewidth} 
         \centering
         \includegraphics[width=\textwidth]{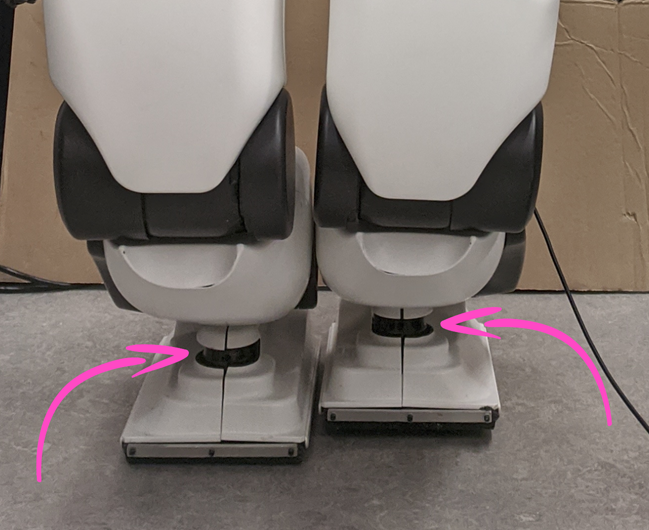}
         \caption{Laser range sensors}
         \label{fig:Proximity sensor}
     \end{subfigure}
     \hfill

    \begin{subfigure}[b]{0.35\linewidth}
         \centering
         \includegraphics[width=\textwidth]{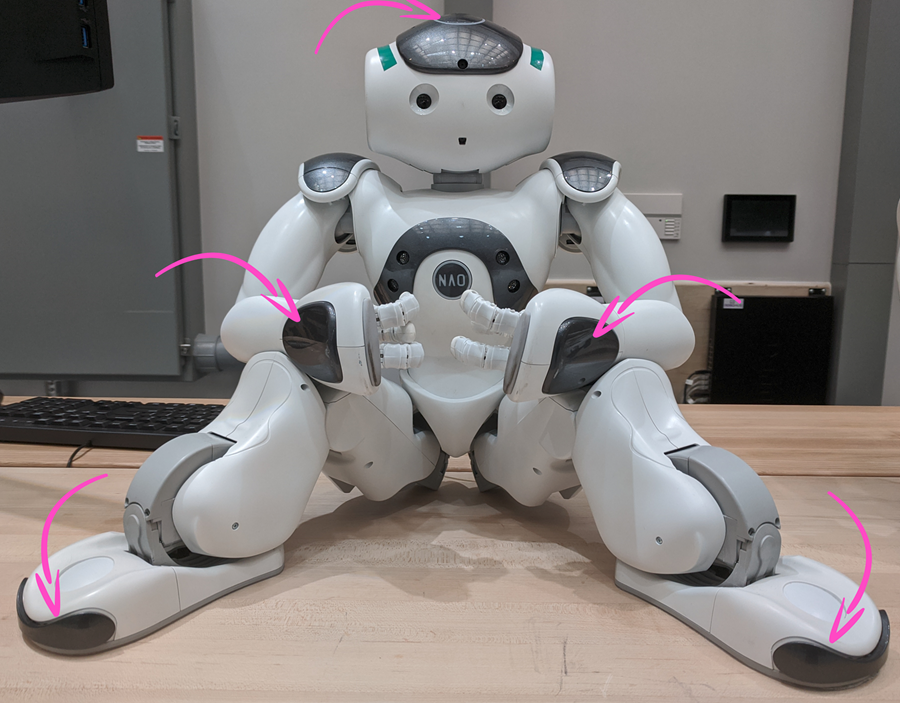}
         \caption{Tactile sensors}
         \label{fig:tactile_sensor_module}
     \end{subfigure}
     \hfill
     \begin{subfigure}[b]{0.28\linewidth}
         \centering
         \includegraphics[width=\textwidth]{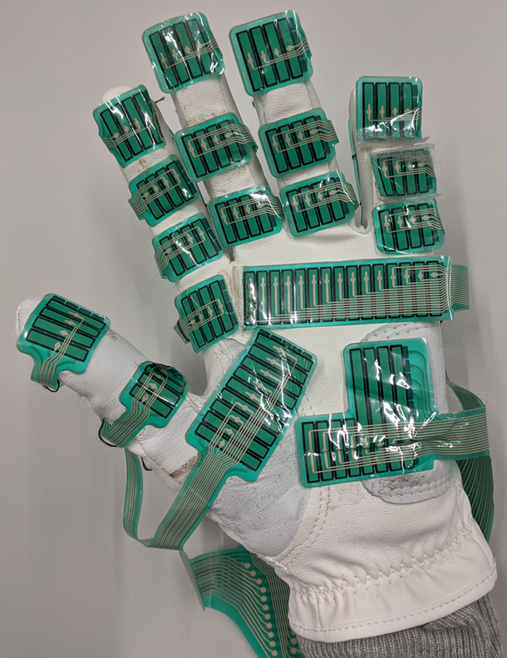}
         \caption{Pressure sensors}
         \label{fig:tactile_sensor_glove}
     \end{subfigure}
     \hfill
     \begin{subfigure}[b]{0.3\linewidth}
         \centering
         \includegraphics[width=\textwidth]{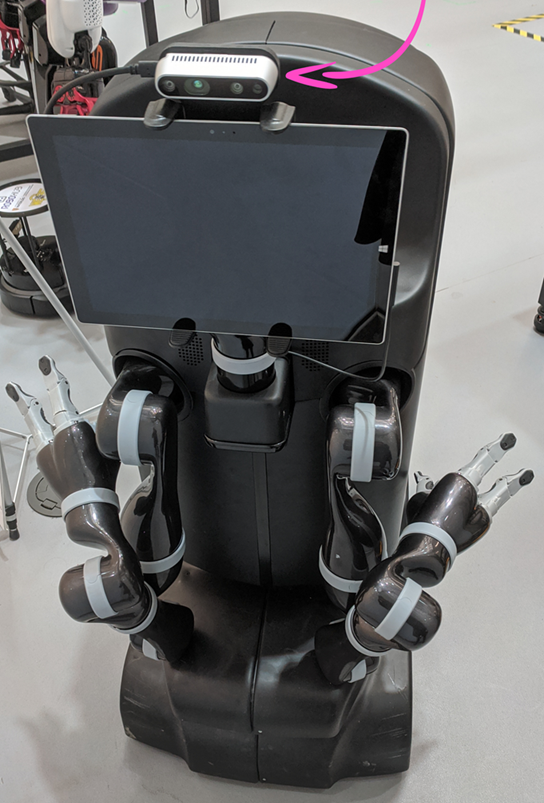}
         \caption{RGB-D sensor}
         \label{fig:camera}
     \end{subfigure}     
        \caption{Sensors commonly employed for pHRI: a)~F/T sensor mounted on the wrist of the REEEM-C robot, b)~RGB-D sensor mounted on the REEM-C head, c)~Laser range sensor embedded in each foot of the REEM-C, d)~Tactile sensors on the NAO robot: capacitive sensors at the head and hands; on/off bumpers on each foot, e)~Flexible resistive pressure sensors mounted on a glove, f)~RGB-D sensor mounted on the MOVO robot.
        }
        \label{fig:tactile, F/T sensors, RGB_D}
\end{figure}

A variety of \textbf{tactile sensors} have been developed and employed specifically for pHRI. On/off tactile sensors have been used to detect human touch or proximity, for example in~\cite{Wong2022.TouchSemantics}, where they allow humans to communicate desired whole-body motions of a humanoid robot through direct pHRI. A Hall effect-based sensing module has been developed to measure forces along 3 axes in~\cite{Holgado2019}
and was shown to measure interaction forces when installed on robot fingertips. The sensor module was later modified to allow force sensitivity adjustments during operation, and augmented with capacitive sensing for proximity measurements~\cite{Holgado2020}, thus also allowing to detect an approaching object before contact occurs.
In~\cite{albini2020pressure}, an artificial skin composed of an array of pressure sensors was used to recognize human hand touch by projecting the 3D pressure distribution onto a 2D image, and classifying hand shapes through machine learning.
A sensor array composed of large flexible 1-D capacitive tactile sensors was developed in~\cite{Leonori2022} to cover the base of a mobile robot, allowing to measure interaction forces while safely engaging in pHRI with the robot base.
Thousands of conformable capacitive sensors disposed in arrays are also used to cover large areas of humanoid robots~\cite{maiolino2013.iCubSkin}. 
For the same purpose, a multi-modal modular artificial skin was introduced in~\cite{GordonCheng.skin}, for which each unit includes proximity, normal force, acceleration and temperature sensing. It has shown multiple times to facilitate pHRI, e.g., in~\cite{GordonCheng.skin, armleder2022interactive}.

Instead of using an array of sensor modules to sense simultaneous contacts in different locations,~\cite{5152595} introduced a flexible sensor sheet embedded into flexible polyurethane that can do just that, while absorbing collisions. Along the same line, the artificial skin introduced in~\cite{Teyssier2021} uses projected mutual capacitance to measure forces applied by multiple contacts, while reproducing the mechanical properties of human skin (Fig.\hspace{0.05cm}\ref{fig:tactile_sensor_module}). With the emergence of wearable sensing, continuous advancements in the development of electronic skin can be expected over the coming years~\cite{wei2022flexible}.

The combination of \textbf{cameras and depth sensors} can produce detailed 3D representations of a robot's surroundings~\cite{9094690}, and can also be used to detect human touch.
Approaches to estimate contact forces without the use of torque or force sensing have also been developed, such as the one introduced in~\cite{Magrini2014.ContactEstimation}. In this paper, contact forces are estimated given (i) joint torques due to contacts estimated based on the robot's dynamic model and given joint positions measured by encoders, as well as the control torques, and (ii) contact locations estimated from images captured by an external depth sensor pointed toward the robot. This approach relies on the concept of the generalized momentum observer~\cite{wahrburg2015cartesian}, calculating change in robot momentum following external disturbances such as contacts with humans.

As \textbf{vision-based tactile sensors} have been materializing for robot manipulation, the concept was extended to a vision-based artificial skin in~\cite{Zhang2020.vision-based-tactile-sensing}: in this design, a camera is used to capture the deformation of a flexible skin covered with a dot grid pattern, thus allowing for 3D force sensing. 

Less ubiquitous, \textbf{proximity sensors}, e.g., LIDAR, detect the presence of nearby entities. However, proximity sensors may not provide detailed enough data to identify what object or human body part is detected, or to predict their motion. 

\textbf{Biometric sensors} are sometimes used in pHRI. For example, electromyography (\textbf{EMG}) sensors capture muscle activity: their signals have been used to estimate interaction forces during pHRI~\cite{su2021deep}, and to modulate compliant controller parameters~\cite{7844516}. Electroencephalography (\textbf{EEG}) sensors capture brain activity in brain-computer interfaces. They have been used to predict motion intentions in assistive robotics, including wearable robotic limbs~\cite{bandara2018noninvasive}.

\textbf{Inertial Measurement Unit }(\textbf{IMU}) sensors typically combine accelerometers, gyroscopes and sometimes magnetometers. They may be installed on a robot to track its motion, or, more often in pHRI applications, they are found in wearable sensors to measure the motion of the human body parts they are placed on. They are therefore commonly used for human motion tracking during pHRI, such as in~\cite{roda2021comparison, campbell2020learning}.

As another option, \textbf{audio sensors} such as microphones capture sound waves and convert them into electrical signals that can be processed by robots.
Taking advantage of AI/ML techniques, 
data captured by audio sensors can be used to 
detect and localize sound sources, parse speech and spoken commands~\cite{bingol2020performing}, or even detect emotions or intents from human voice signals. \cite{9094690} further details the different uses of audio sensors in HRI. Audio sensing can therefore enable humans to communicate with robots through speech or sounds. Audio and visual signals can also be combined for richer human-robot communication, as in~\cite{ashok2022collaborative}.

\subsection{Sensory data integration}

Within the context of HRI, sensor integration, or fusion, refers to the process of combining data from different types of sensors to generate a more comprehensive 
perception of the environment than would be obtained with a single type of sensor. The sensory data integration process in pHRI can be divided into two main components, as suggested in~\cite{9094690}: (1) spatial perception of the environment, task objects and humans, for example fusing visual and depth sensor data, and (2) contact-based perception of the workspace, task objects and humans, for instance fusing F/T, joint torque and tactile sensor data. These two components can then be used separately, or further integrated together. In either case, three levels of data fusion can be defined, following~\cite{gao2022tactile}:

\begin{itemize}
\item Data-level fusion: raw data from multiple complementary sensors is combined. E.g., combining tactile sensor data and visual data from a camera to differentiate human from non-human contacts.

\item Feature-level fusion: features from sensor data are extracted separately, before being combined. E.g., combining human poses and facial expressions extracted from visual data from a camera, with wrenches obtained from F/T sensor readings, to infer human intentions.

\item Decision-level fusion: raw sensor data is processed to extract features that separately result in different outputs of the robot decision process, which are then combined. E.g., combining a target end-effector configuration determined given the pose of an object obtained from camera data, together with desired end-effector displacements determined from wrenches measured by F/T sensors, to generate appropriate robot motions.

\end{itemize}

The reader is referred to~\cite{gao2022tactile}, Table 5, for a schematic representation of each level of data fusion.

In multiple pHRI applications, data integration can be accomplished without explicitly leveraging data fusion for motion control. For instance, human posture estimated from camera data, contact forces by F/T sensors and contact locations by tactile sensors can be directly used to define a variable impedance/admittance controller~\cite{10144527, 7139504, agravante2014collaborative}, a controller combining visual servoing and impedance control~\cite{6906917}, or simply to recognize human activity~\cite{mohammadi2020mixed}.
For an example from the rehabilitation robotics field, EMG data is used in~\cite{1244649, 7926461} to predict human joint torques and adjust parameters of an admittance controller for a gait rehabilitation exoskeleton.

\subsection{System modelling} \label{sub:system modelling perception}

System modelling for pHRI is often centered on interpreting human motions and interaction forces. As exposed in~\cite{al2021improving}, strategies to do so can fall into one or a combination of the following three main categories.
\begin{enumerate}
    \item Model-based strategies: rely on precise mathematical models of the robot and the task,
    \item Human-based strategies: reproduce communication patterns observed in human-human interactions,
    \item Learning-based strategies: leverage AI/ML algorithms to generate models based on data.
\end{enumerate}

\begin{figure}
\centerline{\includegraphics[scale=0.6, clip ]{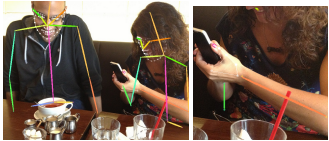}}
\caption{Human body keypoint detection using RGB and depth sensors. Reproduced with permission from~\cite{10000133}.}
\label{fig:human_pose_detection}
\end{figure}

In particular, AI/ML approaches have been shown to contribute toward making pHRI safer and more effective, by enabling human motion identification and prediction~\cite{martinez2019concise}. For instance, multiple algorithms and frameworks are now available to accurately estimate 2D or 3D human pose in real-time~\cite{cao2017realtime, cheng20203d, lee2018propagating, chen2019unsupervised}. Among those frameworks, one of the most widely used is OpenPose~\cite{Cao2021.OpenPose}, an open-source system which can detect the pose of bodies, feet, hands and facial keypoints from multiple people, given 2D visual data; further developments now allow for 3D pose estimation. For example, \cite{10000133} compared the effectiveness of different 2D human pose detection methods and their extension to 3D for close pHRI, with a specific interest in hand detection for pHRI. Fig.~\ref{fig:human_pose_detection}, illustrates an instance of human body keypoint (e.g., head, eyes, arms, hands) detection and 3D pose estimation.

In~\cite{mainprice2013human}, human motion is modelled using a Gaussian mixture model, and then predicted through Gaussian mixture regression. This information is in turned used to predict areas of the robot workspace that will be occupied by a human. In~\cite{10144527}, a supervised ML model is trained to classify intentional and unintentional human touch based on touch location, human posture and gaze direction. For example, in~\cite{fujii2018gaze}, a robotic arm holding a rigid endoscope was developed to be controlled by a surgeon's eye movements, eliminating the need for a camera assistant. Gaze gestures, detected through eye movements, signal the surgeon's intention for camera control. In~\cite{dermy2019multi}, human intention is deduced from gaze direction and physical cues provided through direct manual interaction with the robot.

Relying on human pose detection, real-time human gesture recognition has then been made possible. For example, \cite{Mazhar2018.HandGestures} trained a convolutional neural network (\textbf{CNN}) from hand localization data to detect hand gestures used to communicate actions to the robot. This work was further extended in~\cite{mazhar2019real} to reduce the influence of image background.

Alternatively to vision systems, wearable systems based on IMU can also be leveraged for human posture and gesture recognition. For instance,~\cite{roda2021human} introduces a framework for this purpose, which is further experimentally compared with visual measurements in~\cite{roda2021comparison}. 
Wearable IMU-based sensors have been used in~\cite{Romano2018.CoDyCo} to estimate human internal accelerations and torques when assisting a robot in a direct pHRI scenario. In a scenario where the robot is leading pHRI,~\cite{9447230} models the human based on a digital human model, and training Gaussian processes to predict human postures, given the planned trajectory of the robot and the current human posture measured by a wearable system.

Another common approach to predict interaction forces in pHRI is to use surface EMG signals, such as in~\cite{su2021deep}, where a deep CNN model is trained to output predicted interaction forces, given EMG signals.

Physical interaction data can also be beneficial to human motion and gesture recognition in pHRI. For instance,~\cite{lanini2018human} introduces an approach where a multi-class classifier is trained on human-human interaction data to identify human intention to start, accelerate or stop walking based on motion and interaction force data. A radial basis function neural network is trained in~\cite{liu2019intention} to predict the velocity of a human in contact with the robot at a single point, given the normal contact force, position and velocity of the contact point.

In the absence of tactile sensors, AI/ML approaches have been used to estimate contact forces, such as the radial basis neural network 
proposed in~\cite{8701608}. Authors in~\cite{marban2019recurrent} and~\cite{9594838}, used time series CNN algorithms to detect the forces applied by a surgical tool, based on visual data showing soft-tissue deformation and the deformation of the surgical tool.


\section{Planning} \label{planning}

To ensure safe, efficient pHRI, robot motion planning must include human presence and physical interaction awareness. More specifically, pHRI can be considered as a multi-agent sequential decision problem. Two agents (the human and the robot) select actions based on their given policies, working in coordination with each other to accomplish a shared objective~\cite{vianello2021human}. While subsection~\ref{sub:safety_planning} introduced planning specifically for safety in pHRI, the present section will focus on motion planning approaches for effectively carrying out pHRI.

As discussed above, human awareness can be achieved using different types of sensors, and human motions can be predicted through various approaches leveraging the captured data. Robot motions can then be adapted, given these predictions. 
Typical conventional robot motion planning approaches aim to find feasible paths between initial and final configurations, and avoid obstacles in a known environment. However, to address path planning in complex and dynamic human environments, optimization-based, probabilistic and data driven approaches are more commonly used.
For example, flexibility in planning can be achieved through feedback motion planning, i.e., continuously updating a desired path towards a goal configuration~\cite{6386040}. One approach to do so is through a model predictive control (\textbf{MPC}) framework as proposed in~\cite{li2021hybrid, 8869047}, which enables navigating around dynamic obstacles during the execution of a pHRI task.

Within the context of HRI, one approach to generating robot trajectories may be to mimic human-human interactions. This is for example explored in~\cite{moon2021design}, where human negotiation gestures are reproduced in robot trajectories, in a scenario where human and robot need to settle who grabs an object both are reaching for. However, pHRI may often require planning complex motions. When coupled with complex robot dynamics, \textbf{model-based planning} approaches as described above may require extensive computational resources, compromising the real-time responsiveness of the system. As an alternative, \textbf{learning-based planning} approaches have become more prominent over time.

When it comes to motion planning, learning from demonstration (\textbf{LfD}) often involves pHRI, as synthesized in~\cite{chernova2014robot}. In particular, kinesthetic teaching allows LfD by moving a robot through direct pHRI. A spatiotemporal LfD frameowrk, in which direct pHRI is modeled based on Bayesian interaction primitive, is developed in~\cite{campbell2020learning}: it predicts appropriate robot joint trajectories and contact forces, given the current human pose and forces applied to the robot during social-physical interaction. The concept has been applied to rehabilitation therapy in~\cite{8333285}: a model of the assistive forces applied by the therapist throughout a task is obtained, given interaction force and motion data from kinesthetic teaching demonstrations.
LfD is also used to adapt robot motions to different human partners and new interactions, e.g. as in~\cite{7139393}, where human and robot movements during interactions are correlated by modeling HRI patterns from unlabeled demonstrations, using Gaussian mixture models of interaction primitives.

Movement primitives, as introduced in~\cite{6907265}, are commonly used alongside LfD to represent robot movements or generalize trajectories from demonstrations. The concept is extended in~\cite{lai2022user} to pHRI primitives, which estimate user intent from interaction forces.
Additionally, LfD approaches commonly involve decomposing a complex task into multiple relatively easy sub-tasks, e.g., in~\cite{niekum2012learning}, unstructured demonstrations are segmented into sub-skills, based on dynamic movement primitives and HMMs.
In a similar way,~\cite{7451881} extracted a set of action primitives from a demonstrated sequential task, resulting in a probabilistic representation of the sequence of actions required to complete the task.
The authors in \cite{8607032} employed the concept of dynamic movement primitives to characterize the interaction and learn the robot's trajectories in real-time through contact-based pHRI.

Inverse reinforcement learning (\textbf{IRL}), or intention learning, is a subset of LfD approaches in which the robot, instead of learning demonstrated motions, learns a reward function. In particular,~\cite{9340865} leverages this concept for the kinesthetic teaching of a mobile robot: given direct pHRI forces, IRL is used to adapt parameters of a navigation cost function, ultimately correcting robot trajectories. Similarly, collaborative arm trajectories are adapted online through IRL in~\cite{losey2022physical}, where appropriate corrections to the objective function are learned, given interaction forces.


\section{Control} \label{control}
The choice of appropriate control approaches is a key factor in the safety and effectiveness of pHRI applications. When a human and a robot are closely working together, task execution and control is likely to be shared between them (e.g. the human may be controlling end-effector position, while a robot controller maintains its orientation), leading to the concept of \textbf{shared control}. An overview of the topic oriented towards pHRI can be found in~\cite{losey2018review}, where central themes include the division of control and communication between human and robot. Integrating latest AI/ML techniques allows for greater flexibility in shared control of pHRI, as suggested in a recent survey~\cite{9501975}. These advancements have paved the way towards the concept of \textbf{shared autonomy}, where robots can be programmed to dynamically adapt their autonomy level in function of context. For example, a robot can have a lower level of autonomy when performing tasks that require human input, and higher autonomy for repetitive tasks.

Controlling robot motions to be compliant to physical interactions is also a common approach in pHRI, as documented in~\cite{khan2014compliance}. While control techniques for safety have been discussed in~\ref{sub:safety_control}, this section discusses recent advancements in control adapted to direct pHRI, with a focus on compliant control for collaborative robotic arms. Inherent instability in robots like humanoids and ballbots necessitates additional attention to stability when engaging in contact-based pHRI, a topic that will also be addressed in this section.

Compliance in a robot can typically be achieved through:
\textbf{Passive compliance:} 1) leveraging inherent flexibility in the mechanical structure of the robot and 2) employing compliant actuators, e.g. series-elastics or pneumatic actuators.

\textbf{Active compliance:} defining control laws modulating the resistance of the robot to external forces or torques, such as admittance and impedance control laws.

Utilizing compliant control (active compliance) allows robot actions to be adjusted in response to external forces, thus promoting safety and mitigating the potential for injury~\cite{li2020control, khan2014compliance}. With its versatility, this software-driven method can be implemented across diverse applications, bolstering safety measures and facilitating effective interaction in scenarios involving pHRI.

The mechanical impedance of a structure (such as the body of a robot) relates the force acting on it to its displacement (or velocity), indicating how the structure resists external forces~\cite{sharifi2014nonlinear}. Put simply, it is the ratio of the displacement at the specific point on the robot where the force is applied by the magnitude of the applied force. 
Conversely, admittance, the inverse of impedance, is the ratio of applied force to displacement at that point. Admittance takes force as input and provides displacement (or velocity)
~\cite{rhee2023hybrid}. It is noteworthy that these concepts can be described in either Cartesian space or joint space.

The parameters of mechanical impedance/admittance (inertia, stiffness, and damping) serve as metrics to measure a robotic system's resistance to motion when subjected to force by a human, such as at its end-effector.
In other words, in compliant control, the robot's force/motion in response to external motion/force is controlled by representing the robot as an impedance/admittance element and fine-tuning its parameters. In recent years, these types of controllers have risen as notably efficient techniques in the pHRI realm.

For example, admittance control is used in~\cite{haninger2022model} to generate safe robot motion control during pHRI, while robot trajectories are obtained with a model predictive control approach that considers the progression of a predefined task. Instead, impedance control is used in~\cite{7934303} to ensure compliance to external interaction, while reinforcement learning is used to identify if a human is leading or following in the completion of a task requiring pHRI. Authors in~\cite{brahmi2019compliant} proposed an adaptive tracking controller that relies on the modified function approximation technique to estimate the uncertain dynamics of an exoskeleton robot and reach a compliant control in a contact-based pHRI.

One of the main research directions in compliant control for pHRI is in modulating the stiffness, damping and inertia of the robot during operation, i.e., \textbf{variable impedance (VI) and variable admittance (VA) control} schemes. In particular, \cite{keemink2018admittance} provides a comprehensive discussion of admittance control for pHRI and its distinction from impedance control. Since the concept was introduced in~\cite{huang1992compliant}, strategies 
have been developed based on various control inputs and feedback as reported in~\cite{9325872}, and further synthesized in Table~\ref{tab:variable_strategies}.

\begingroup

\renewcommand{\arraystretch}{1.5} 

\begin{table}[!ht]
    \centering
    \begin{tabular}{p{0.22\linewidth} | p{0.65\linewidth}}
        \textbf{Approach} & \textbf{Control commands generated based on} \\
        \hline
        Velocity-based & joint, end-effector, or base velocity \\
        Force-based & wrenches at the end-effector or other interaction forces on the body of the robot \\
        EMG-based & EMG signals from human operator muscles \\
        Stability-based & stability margins of the controller, given interaction dynamics  
    \end{tabular}
    \caption{Strategies for admittance or impedance variation} 
    \label{tab:variable_strategies}
\end{table}

\endgroup

Various methods for adjusting VI and VA control parameters, such as adaptive control and other conventional control strategies, as well as stabilization approaches, are addressed in subsection~\ref{sub:tradiditional control}.

Often, robots interacting with humans require more intelligent behavior than passively following interaction forces. AI/ML algorithms make it possible to handle a range of pHRI scenarios by programming complex robot behaviours, given data from sensors measuring pHRI~\cite{semeraro2023human}. 
The next subsections will cover how AI/ML algorithms have been applied to VA and VI control in pHRI applications. We focus in particular on more artificial neural networks in~\ref{sub:ANN control}, deep learning in~\ref{sub:deep learning control}, reinforcement learning in~\ref{sub:RL control}, and learning from demonstration in~\ref{sub:LfD control}.

\subsection{Traditional control} \label{sub:tradiditional control}

A VA control approach is introduced in~\cite{li2020control}, combining a force sensor and a force observer to safely allow a human to physically guide a robot making contact with objects of unknown stiffness.
Several VA control laws for direct pHRI were explored in~\cite{chen2022human}, relying on interaction forces and velocity for control parameter modulation. The control laws also ensure passivity through power envelope regulation, which bounds within a safe range the power injected into the system by interaction forces.

In~\cite{lee2022real}, the argument is made that when a robot is controlled with an admittance control law, a force feedback is generated during direct pHRI by a combination of the robot motion and the human hand impedance, which distorts the measured interaction forces. The paper therefore introduces a variable hand impedance compensation scheme. Instead,~\cite{yu2020simplified} argues that since admittance controllers rely on the inverse kinematics or the Jacobian, they require a highly accurate model of the robot for precision. As an alternative, the paper introduces admittance control methods based on end-effector orientations.

As for impedance control, a velocity-based VI controller, with proven stability and convergence given uncertain contact impedance characteristics, is proposed in ~\cite{dong2020physical}. The problem of model uncertainty is addressed in~\cite{sharifi2014nonlinear}, with the introduction of four different model reference adaptive impedance controllers.

When addressing the control of inherently unstable robots in pHRI, the control challenge is broadened by accounting for the human's present state, predicting their future state, and preserving stability through concurrent motion and contact control. Predicting the human's future state entails estimating their motion intentions. Essentially, the issue in pHRI with unstable robots lies in determining how to anticipate the human's intentions to develop robot controllers that are responsive to the human's movements (human's transition to future states) while attempting to maintain stability amid interactions influenced by both human forces and the dynamics of the robot itself~\cite{8093992}.

This multitask optimization challenge is often addressed using Quadratic Programming (\textbf{QP}), which minimizes the disparity between actual and reference task values. QP formulation allows for adaptable control solutions, covering inverse kinematics, inverse dynamics, and momentum-based challenges in either position or torque-controlled robots~\cite{vianello2021human}. QP serves as a method for defining a constrained, at times hierachical, optimization problem, often formulated to find the control inputs that minimize the tracking error on one or multiple tasks. The constraints are critical to ensure safe robot behavior, such as joint position, velocity, and torque limits, as well as limits on allowable contact wrenches to maintain balance by keeping the ZMP within the support polygon, and in somple implementations, also imposing center of mass (\textbf{CoM}) trajectories~\cite{8463167, 8624995}.

The interaction forces exerted between human and humanoid robot can be reflected in the QP constraints as part of a dynamic robot model subjected to external forces. For example,~\cite{8463167} implemented a QP controller for human-humanoid physical collaboration with the addition of constraints for contacts and collision-avoidance. To achieve compliance during the interaction, a stack of tasks can be defined in a hierarchical QP controller. These tasks can for example include position regulation based on human ergonomics, 
interaction force tracking, or motion tracking~\cite{tassi2022adaptive, tassi2023multi}. 
The interaction dynamics between humans and humanoid robots (the forces exchanged between them) can be incorporated into QP constraints, modeling the robot's dynamic response to external forces.

Beyond a traditional QP, inherently unstable robots can greatly benefit from a stabilizer to maintain balance during physical interaction tasks, as introduced in subsection~\ref{subsub:stabilization}. Within this context, a few comprehensive approaches to whole-body compliant control have been introduced. In~\cite{hoffman2018whole}, a controller is proposed, comprising a compliant stabilizer based on the zero moment point (\textbf{ZMP}), measured via force/torque sensors located at the robot's ankles, to complement a whole-body inverse kinematics
engine. In~\cite{8793258} instead, the robot is equipped with a tactile sensor skin to capture interaction forces
and employs a PID controller for balancing through ZMP tracking.

In~\cite{7803289}, a wheeled humanoid robot is made to guide physical interactions during dance training by controlling the height of its COM.
In this approach, the robot engages in a dance with its human partner by adhering to a predefined trajectory
while compliantly responding to the partner's movements.
This is facilitated through impedance control, which determines the robot's joint accelerations according to the forces exchanged between the robot and the human partner. The robot is also programmed to convey its direction of motion with CoM height variations.

\subsection{Artificial neural network} \label{sub:ANN control}

Different solutions based on artificial neural networks (\textbf{ANN}) have been proposed for variable impedance and admittance control.
In~\cite{pugach2016touch}, an ANN is trained to output the force admittance of a robotic arm, in function of interaction forces measured by tactile sensors installed on the arm.
An admittance control approach is proposed in~\cite{8884133}, where an ANN is set up to feedback linearize the robot dynamics during pHRI, and Lyapunov stability analysis is leveraged to obtain ANN weight tuning laws. 
To enhance cooperation during pHRI,~\cite{app11125459} introduced an admittance controller based on a combination of hedge algebras and ANN, which are used as an alternative to dynamic admittance model identification.

Controlling compliant robots, whether through active compliance or tracking control, in the presence of joint flexibility poses challenges due to uncertainties in dynamics. Employing ANNs can effectively address these challenges by handling the complexities of flexibility dynamics.

In~\cite{9548781}, an adaptive impedance controller is introduced for human-robot co-transportation tasks. This controller, incorporating an admittance-based radial basis function NN,
error constraints, and input constraints, enables the tracking of human hand position and interaction force through vision and force sensing. 
In~\cite{10173753}, the precise tracking of flexible 
robot joints in uncertain environments is addressed using a Lyapunov-stable adaptive neural network controller. The controller comprises two loops: a force-based outer loop and a position-based inner loop. The outer loop generates the reference trajectory using interaction force error and estimated environment stiffness, while the inner loop focuses on accurate position tracking with neural network compensation for uncertainties.

\subsection{Deep learning} \label{sub:deep learning control}

Deep learning can be instrumental in advancing the field of pHRI by empowering robots to perceive, plan and respond to human intentions with greater accuracy and adaptability. Through deep neural networks (\textbf{DNN}s) such as deep ANNs, CNNs or recurrent neural networks (\textbf{RNN}s), complex mappings can be achieved between sensory inputs, such as visual, tactile, joint encoder, and F/T sensor feedback, and appropriate motor responses. Given the intricate nature of pHRI dynamics, neural networks require significant complexity, with a high number of layers, abundant data, and appropriate regularization. This is where deep learning methods prove to be highly advantageous.

DNNs find multiple applications across a perception-planning-control system. Literature in which DNN techniques are used for perception and planning, such as~\cite{marban2019recurrent,9594838, losey2022physical} have been previously discussed in the Perception section~(\ref{perception}), under its system modelling subsection (\ref{perception}.\ref{sub:system modelling perception}) and planning section (~\ref{planning}).
DNNs have been used to interpret subtle human cues, anticipate movements, and adjust robot behaviour accordingly, towards fostering more natural and intuitive human-robot interactions~\cite{9666912, su2021deep}.
Moreover, deep learning has been applied to facilitating the development of control strategies involving
learning from human demonstrations, adapting to dynamic environments, and optimizing robot actions to ensure safety and efficiency. This will be covered in the following two subsections, which will include the application of deep learning to RL and LfD.

\subsection{Reinforcement learning} \label{sub:RL control}

Reinforcement learning (\textbf{RL}) can be employed to model complex task dynamics, which in turn can be used to optimize VI or VA control parameters as proposed in~\cite{roveda2020model}. In this paper, the control parameters of a VI controller are optimized online through MPC, given the objective to minimize human effort during interaction and a model of HRI dynamics generated using ANNs. 
RL is used in~\cite{9786637} to automatically infer VI parameters of a robotic knee prosthesis by mimicking the motion of the intact knee.
Instead, RL in~\cite{7353494} is employed to learn the damping coefficients of a VA controller that minimize jerk in point-to-point movements during co-manipulation tasks.

Combining the deep deterministic policy gradient algorithm (\textbf{DDPG})~\cite{lillicrap2015continuous} 
and reward function optimization is proposed by~\cite{liu2021deep} for safe human-robot collaboration in the manufacturing context: in this paper, a reward function is optimized to effectively learn collision avoidance policies. Instead, within the context of rehabilitation robotics,~\cite{9849514} introduced a controller for a robotic orthosis employing a two-stage deep RL strategy. 
Firstly, optimal human gaits are learned using deep RL-based imitation learning of a healthy human model. Then, models of weakened soleus muscles are developed and used to train a robotic orthosis policy for walking assistance.

\subsection{Learning from demonstration} \label{sub:LfD control}

Being based on human demonstrations, LfD is potentially more suitable than the typical RL approach of trial-and-error exploration, in the case of pHRI applications in which random exploration could be unsafe~\cite{9531394}. For instance, in~\cite{6224877}, a teaching interface based on learning from demonstration is used to decrease the stiffness of an impedance-controlled robot in the Cartesian space, given displacements of the end-effector generated by direct pHRI. This work was later extended in~\cite{kronander2013learning} with an interface to increase stiffness based on measured interaction forces, and a mechanism to modulate stiffness either in the Cartesian or the joint space. 

Adopting a different strategy,~\cite{9842322} combines IRL and RL to tune impedance control parameters in an optimization framework where cost functions for pHRI performance are obtained from IRL, and then used to determine impedance parameters through RL.
Another LfD approached introduced in~\cite{8794065}, a deep RL algorithm is used to train control policies through a customizable multi-objective reward function derived from motion capture data of human-human handshakes and hand claps.

The diversity of human behaviours drives the complexity of pHRI. The development of \textit{intention-oriented} control systems is required, in order to generate appropriate robot motions, given human intentions inferred from a combination of sensors. Further data-driven control approaches have yet to be extensively researched to make this a reality.


\section{Computational Enhancement} \label{computational_enhancement}
Contact-based pHRI demands rapid response and real-time performance from robots. When AI/ML is involved, the dynamic nature of the problem, coupled with changing interaction forces and human states, often necessitates resource-intensive deep learning methods. Consequently, computational constraints pose significant bottlenecks in the system, particularly with algorithms like fully connected neural networks, CNNs and RNNs~\cite{thompson2020computational, baskakov2021computational}.

In this context, this section addresses computational challenges from both a hardware and a software perspective. The following section focuses on hardware technologies aimed at accelerating computation with a focus on processing units, while the subsequent subsection discusses control architectures and paradigms for achieving real-time and fast perception-planning-control systems.

\subsection{Hardware architecture} \label{hardware architecture}

To address computational challenges, hardware and software system modification represent one approach to reducing computation time. For instance, various processor units and specialized electronic boards have been proposed to expedite computations. Incorporating graphics processing units (\textbf{GPUs})~\cite{dally2021evolution}, tensor processing units (\textbf{TPUs})~\cite{9774600}, and field-programmable gate arrays (\textbf{FPGAs})~\cite{1708640} in conjunction with conventional central processing units (\textbf{CPUs}) has rendered deep learning algorithms viable for high-performance computing.

GPUs excel in parallel processing, making them advantageous for training complex neural networks and analyzing sensor data quickly for real-time decisions in robotics. TPUs, on the other hand, are specialized hardware accelerators designed for machine learning tasks, particularly deep neural networks. They excel in matrix calculations and are preferred for both training and inference processes in robotics, especially for rapid data throughput and minimal latency. For their part, FPGAs offer adaptability and are useful for instant processing of sensor data, swift implementation of control algorithms with minimal delay, and accelerating machine learning computations within robotic frameworks. 

A study in~\cite{9350173} evaluated the performance of CPUs, GPUs, and FPGAs in solving the forward dynamics of articulated robotic arms. This task involves spatial algebra and the derivative of the Recursive Newton-Euler Algorithm. The study underscored the importance of this comparison, given that computing the gradient of rigid body dynamics typically consumes 30\% to 90\% of total computational time in nonlinear MPC implementations.
The results revealed that the GPU and FPGA implementations completed the forward dynamics solution three times faster than their CPU counterpart, thanks to more efficient utilization of parallelism and customization.

To accelerate deep learning models, parallelizing computations is key. This involves dividing data or models into smaller chunks and processing them concurrently across multiple devices. By harnessing the power of parallel processing, such as utilizing GPUs alongside FPGAs or leveraging multiple CPU cores, processing time is significantly reduced. For instance, in~\cite{tu2019power}, a power-efficient implementation of DNN was suggested for both FPGAs and GPUs to accelerate the DNN computations, comprising both a CNN block and a fully connected NN block. By allocating the CNN part to the GPU and utilizing the FPGA for the fully connected layers, both could be processed in parallel. Through model breakdown and distributed processing, swifter computation could be achieved, along with a significant reduction of power consumption.

\subsection{Software architecture} \label{software architecture}

Delving into robotic system architecture, the complexities inherent in designing systems capable of interacting with dynamic real-world environments, including humans, call to be explored. At its core, a robotic system is a complex communication network between sensors and actuators, geared towards accomplishing a defined set of tasks. However, the variability and uncertainty of pHRI scenarios, coupled with the diverse array of sensors and actuators, pose a level of intricacy that demands meticulous design and practical implementation strategies.
Optimal architecture for rapid performance is also crucial for human safety, i.e., real-time and fast sensing, planning, and acting is important to ensure safety in a pHRI scenario.

A robotic system architecture can be described with two key aspects: structure and style. The structure concerns the way in which the system is broken down into manageable interconnected subsystems, whereas the style involves the computational framework that defines communication among components within the architecture~\cite{kortenkamp2016robotic}. While there may not be a one-size-fits-all architecture for robotic systems, certain paradigms have emerged as valuable design frameworks. Below are listed several prominent architectural structures:

\textbf{Deliberative architecture (Sense-Plan-Act):} 
This architecture, which was among the first proposed architectures, comprises three core subsystems: sensing, planning, and execution, arranged sequentially in a hierarchy. Sensor data is relayed to the planner, which then communicates with the controller to issue actuator commands. Deliberative architectures are flexible, scalable, and intelligent 
due to their ability to process sensory information in the planning module and to make decisions on actions.
However, they come with a number of drawbacks: the planning stage often slows down the controller due to computational limitations, and the controller's lack of direct sensor access hampers system reactivity~\cite{ingrand2017deliberation}. 
The nested hierarchical controller~\cite{4048968} and the US National Institute of Standards and Technology (NIST) real-time control system~\cite{albus1995nist} are two examples of this structure.

\textbf{Reactive architecture (Sense-Act):} Reactive architectures operate under the premise that a robotic system can react to sensor inputs without requiring internal representation of 
sensory data (information) or planning. Such architectures comprise basic rules or behaviours that prompt actions in response to stimuli. While reactive architectures offer speed, robustness, and straightforward implementation, they often sacrifice flexibility, scalability, and intelligence.
Since there is no planning module to process perceived sensory data and make decisions for actions, the sensory data and actions are confined to those that are hard-coded
~\cite{ingrand2017deliberation}. They are suitable for straightforward robotic systems operating in consistent, foreseeable conditions. And example of a suitable scenario could be basic obstacle avoidance, where a mobile robot is programmed to move along a straight path, and to simply shift to the left or right upon sensing an obstacle.

\textbf{Hybrid architecture:} The principle behind hybrid architectures in robotics is to combine reactive and deliberative elements, in order to leverage their respective strengths into different levels of control. For instance, low-level (behavioural) actions may be handled reactively, while high-level planning may be handled deliberatively~\cite{arkin1998behavior, qureshi2004cognitive}. 
In this context, low-level control is primarily concerned with executing localized, short-term behaviours at the sensor and actuator level (such as commanding the robot to turn left or right to avoid an obstacle). An executive, intermediate level oversees the translation of high-level plans into actionable low-level behaviours and manages exceptions (such as navigating to a destination). High-level planning then involves deliberate decision-making and long-term strategizing to optimize robot behaviour (such as planning to reach a destination and perform a given task). Although hybrid architectures provide a blend of the advantages found in both reactive and deliberative architectures, such as adaptability, modularity, and robustness, they may present challenges in terms of their design, implementation, and debugging processes due to the complexity of this architecture.

\textbf{Subsumption architecture:} As proposed by~\cite{1087032}, the subsumption architecture presents a real-time control option, as an alternative to the sense-plan-act paradigm. In this architecture, higher-level behaviours exert control over lower-level ones, facilitating the delegation of minor tasks to lower levels. Hence, it is designated as a \textbf{behaviour-based} structure. Each behaviour, such as map-building, exploration, wandering, and obstacle avoidance, is realized as a layer of finite state machines interconnected with sensors and actuators. This architecture is designed to allow multiple behaviours to be evaluated simultaneously and activated sequentially, through an arbitration mechanism determining the prioritized hierarchy of behaviours in real-time.
 
The subsumption architecture has proven to be highly effective through numerous implementations, such as presented in~\cite{horswill1993polly, mataric2018integration}
The robots programmed under this architecture have been shown to exhibit real-time performance and responsiveness, due to their ability to continuously perceive and respond to changes in their surroundings, thus indicating the subsumption architecture to be well-suited for dynamic or human environments. Consequently, it is a promising option for scenarios involving pHRI, where real-time sense-plan-action capabilities are crucial, and where traditional AI may fail to provide sufficient response speed.

Traditional AI approaches divide tasks into intricate subsystems such as perception, modeling, planning, execution, and control, which are carried out sequentially. 
Each of these subsystems may entail complexity and consume significant time. In contrast, the subsumption approach simplifies control by organizing tasks into parallel 
layers, each representing a specific behaviour. Each layer can independently control the robot in a basic manner, thus operating more swiftly compared to traditional AI methods~\cite{toal1996subsumption}.
However, while this architecture is highly reactive, it currently lacks effective long-term planning or behaviour optimization capabilities, thus posing challenges when it comes to achieving long-term objectives~\cite{kortenkamp2016robotic}. 

Aside from determining the architecture of a robotic system, selecting its style can be just as crucial. The style influences how the different components of a system, such as the planner, controller, and sensors, interact with each other. This communication is typically facilitated by middleware, which can for example take the format of client-server or that of publisher-subscriber. Client-server middleware, such as remote procedure call (RPC)~\cite{amoretti2010architectural} involves clients sending requests to a server, risking deadlocks (which may for example occur due to server crash). In contrast, publisher-subscriber middleware broadcasts messages asynchronously, such that the control flow isn't tied to any specific order, thus reducing the impact of missing or out-of-order messages.~\cite{chavan2015review}. Robot Operating System (ROS) is a popular robotics middleware, which has drawn significant attention from research and industrial communities since its initial release in 2007~\cite{quigley2009ros}. It is primarily built on the publisher-subscriber style of communication, although it also incorporates support for the client-server style.

In the past sections, the focus has been mostly on the technical functionality and performance of robotic systems that directly physically interact with people. In contrast, the next section will delve into the human experience of pHRI. 


\section{Ethics} \label{ethics}
\begingroup

\renewcommand{\arraystretch}{1.5}

\begin{table*}[!ht]
\begin{tabular}{ p{0.45\linewidth} p{0.45\linewidth} }
    \textbf{Ethical issues} 
    & \textbf{Potential strategies} \\

    \hline
    \rowcolor{Gray}
    Preventing harm to those who interact with robots, including physical, social and emotional impact of robots
       & Safety guidelines for robot design~\cite{martinetti2021redefining}, protocols for robots touching humans, consider human emotional needs and the formation of emotional bonds in design~\cite{riek2014codeOfEthics} \\

    Ensuring users understand robot capabilities and intentions 
        & Transparency~\cite{9162045} and explainability measures~\cite{setchi2020explainable} \\

    \rowcolor{Gray}
    Protecting user privacy
        & Cybersecurity approaches~\cite{lera2017cybersecurity}, define and follow 
        privacy laws and regulations~\cite{riek2014codeOfEthics}\\

    Develop appropriate levels of trust in users
        & Predictability and communicativity~\cite{Hamacher2016TrustpHRI}\\

    \rowcolor{Gray}
    Ensure user comfort, well-being and autonomy
        & Predictability~\cite{9698842}, interaction design and transparency~\cite{fronemann2022EthicsUX} \\

    Eliminate biases and discrimination
        & Community engagement, transparency and explainability measures~\cite{howard2018bias},
            avoid explicit computational evaluation of identity characteristics~\cite{williams2023eye},
            alternative robot morphologies~\cite{riek2014codeOfEthics} \\

    \rowcolor{Gray}
    Maintain human dignity and ensure fair labor practices
        & Support worker training and participation, transparency~\cite{meissner2020friend}
\end{tabular}
    \caption{Imminent questions that need to be addressed regarding pHRI, along with potential solutions to explore}
    \label{tab:ethical_strategies}
\end{table*}
\endgroup

When talking about deploying pHRI in the real world, it is critical to address questions of ethics encompassing physical and psychological safety, transparency, privacy and work environment concerns~\cite{van2022ethical, etemad2022ethical, moon2021ethics}. Some of the questions needing most urgent consideration for the responsible and beneficial integration of robotic technology in society are included in Table~\ref{tab:ethical_strategies}.

While solutions have yet to be implemented, tested and evaluated in the context of pHRI, and many questions have yet to be answered, e.g., what safety measures ensure psychological safety during pHRI? How does one make robot programming transparent? How do biases show in pHRI? Who will be directly and indirectly affected by pHRI applications? How will major technology companies or governments engage on future employment concerns? 

As part of the solution, a code of HRI ethics proposed in~\cite{riek2014codeOfEthics} includes physically assistive robots. Ethical issues that influence the intention of people to use 
interactive robots have been investigated in~\cite{etemad2022ethical}, leading to recommendations for robot design. 
Additionally, the integration of established usability and user experience design principles into social HRI and collaborative robots is investigated in~\cite{fronemann2022should}, towards ensuring human comfort and well-being. 

As research on the ethics of HRI is gradually emerging, we can look forward to further developments in the ethics of pHRI. However, the currently limited literature documenting problems and solutions relevant to contact-based pHRI indicates a need for researchers to dedicate resources to this problem, before robots become more integrated into everyday life. One significant challenge that was not explicitly mentioned yet but requires pressing attention, is that of comprehending how working in close proximity to robots impacts humans. Both on the short or on the long term, gaining a deeper understanding of robots' influence and their implications on society is critical to ensuring the well-being of future populations across the globe (and beyond).

\begin{figure*}
     \centering
     \includegraphics[scale=0.65, clip]{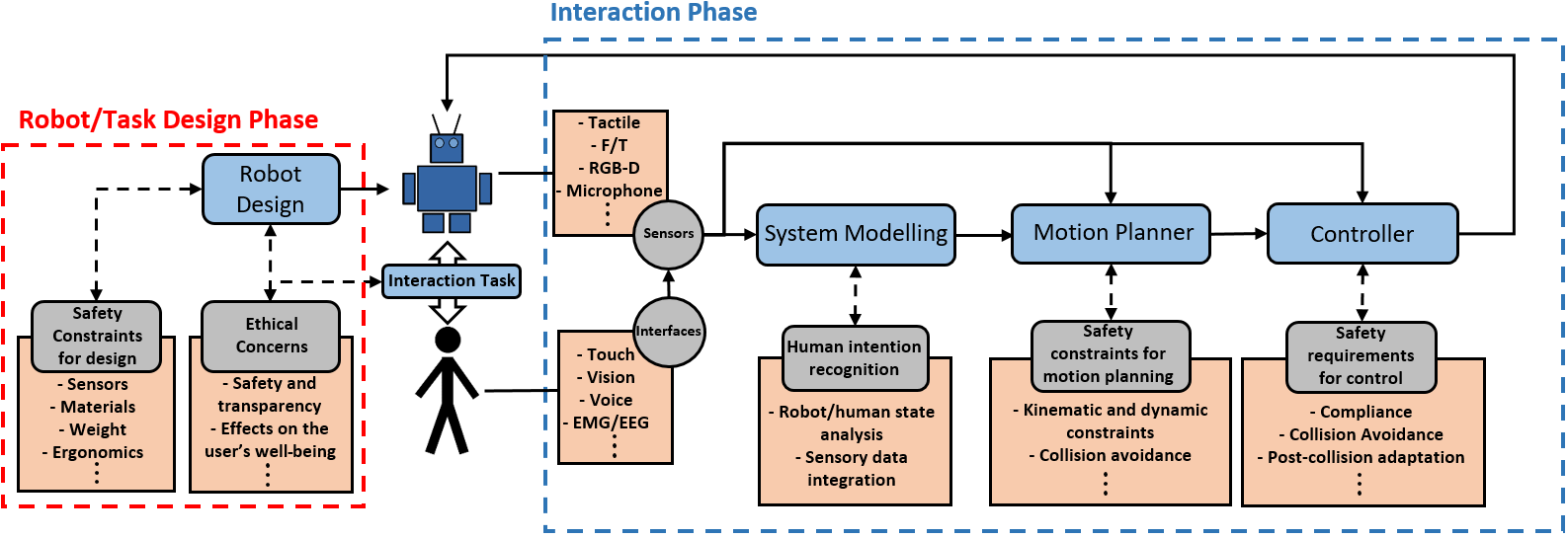}
     \caption{Schematic illustration of a typical pHRI framework, demonstrating the interconnections between various modules. Emphasizing safety as a critical aspect, the figure highlights the overarching need to address safety at each step, as well as to consider ethics from the start.
     } 
     \label{fig:pHRI_framework}
\end{figure*}


\section{Conclusion} \label{conclusion}
The future of pHRI holds promising opportunities, with the potential to revolutionize industry and everyday life.
However, significant challenges have yet to be tackled: the analysis above leads us to conclude that robotics, and pHRI in particular, is still in its infancy. 

In this survey, we have covered diverse categories of pHRI with a specific emphasis on contact-based interactions and explored various interconnected aspects of making the interaction safe and effective.
To leave the reader with a sense of the interconnection between each of the aspects covered above, Fig.~\ref{fig:pHRI_framework} provides a schematic representation of the collaboration and communication between ethics to design, planning and control. We propose to end with an outline of the challenges we have identified
as most urgently requiring further investigation and attention:

\subsection{Design} The design and testing of robots for pHRI require substantial engineering efforts and time investment due to constraints including use of materials, battery power, sensor and actuator availability, as well as robot functionality, locomotion abilities, and user-friendliness. Currently, robot utilization remains limited to specific scenarios and targeted tasks, primarily within industrial, rehabilitation and medical robotics. The development of versatile robots capable of seamlessly integrating into various aspects of human life remains an ongoing challenge.  

\subsection{Perception} It is crucial for robots to accurately perceive the dynamically changing environment and humans around them.
To achieve comprehensive perception and develop socially aware robots, employing multi-modal sensory systems becomes inevitable, for example combining vision, touch, and audio signals.
With advancements in AI/ML algorithms and their integration into robotics, e.g., the latest trend to integrate a generative pre-trained transformer into robot communication systems,
there is great potential to achieve natural and intuitive interactions.

\subsection{Adaptability} Developing robots that can adapt to diverse human partners, environments, and tasks still needs significant effort. With currently trending approaches, this would require large AI/ML models and datasets, long learning periods and substantial computational resources, while still having to deal with limitations in the design of robots.

\subsection{Safety and compliance} Prioritizing human safety, for instance by ensuring\linebreak robot compliance to interactions and adequate response to human actions, requires a tight integration of sensors, planning and control systems.
Additionally, to properly ensure psychological and social safety in pHRI scenarios, further research on the impacts of robots on humans within pHRI contexts is acutely called for. Developing robots that exhibit intelligent behaviour and ensuring that people place an appropriate amount of trust in robots remains a persistent challenge.

\subsection{Ethics} It is becoming essential for engineers, researchers, industries and governments to prioritize ethical concerns regarding HRI and pHRI. Open conversations and extensive research needs to be conducted to explore effective ways of addressing the ethical challenges identified in the previous section, and the new challenges that have yet to be uncovered. Ultimately, this will hopefully lead to developing robots that work in the interests of people, and to increasing the interest of people in employing robots that work alongside them. In-depth ethical studies are prompted to ensure safety in pHRI, address the concerns related to robots replacing humans, to appropriately consider cultural and social differences in HRI, among all the significant ethical challenges that remain open. 

Our hope is that by engaging with these challenges on all fronts, robotic technology will be developed to realize the promises it holds, while enabling future society to thrive.

\bibliographystyle{unsrt}  
\bibliography{references}  

\begin{thebibliography}{100}

\bibitem{rawassizadeh2019manifestation}
Reza Rawassizadeh, Taylan Sen, Sunny~Jung Kim, Christian Meurisch, Hamidreza Keshavarz, Max M{\"u}hlh{\"a}user, and Michael Pazzani.
\newblock Manifestation of virtual assistants and robots into daily life: Vision and challenges.
\newblock {\em CCF Transactions on Pervasive Computing and Interaction}, 1:163--174, 2019.

\bibitem{henschel2021makes}
Anna Henschel, Guy Laban, and Emily~S Cross.
\newblock What makes a robot social? a review of social robots from science fiction to a home or hospital near you.
\newblock {\em Current Robotics Reports}, 2:9--19, 2021.

\bibitem{de2008atlas}
Agostino De~Santis, Bruno Siciliano, Alessandro De~Luca, and Antonio Bicchi.
\newblock An atlas of physical human--robot interaction.
\newblock {\em Mechanism and Machine Theory}, 43(3):253--270, 2008.

\bibitem{castro2021trends}
Afonso Castro, Filipe Silva, and Vitor Santos.
\newblock Trends of human-robot collaboration in industry contexts: Handover, learning, and metrics.
\newblock {\em Sensors}, 21(12):4113, 2021.

\bibitem{ben2018robots}
Mordechai Ben-Ari, Francesco Mondada, Mordechai Ben-Ari, and Francesco Mondada.
\newblock Robots and their applications.
\newblock {\em Elements of robotics}, pages 1--20, 2018.

\bibitem{evjemo2020trends}
Linn~D Evjemo, Tone Gjerstad, Esten~I Gr{\o}tli, and Gabor Sziebig.
\newblock Trends in smart manufacturing: Role of humans and industrial robots in smart factories.
\newblock {\em Current Robotics Reports}, 1:35--41, 2020.

\bibitem{9024658}
Khalid~Hasan Tantawi, Alexandr Sokolov, and Omar Tantawi.
\newblock Advances in industrial robotics: From industry 3.0 automation to industry 4.0 collaboration.
\newblock In {\em 2019 4th Technology Innovation Management and Engineering Science International Conference (TIMES-iCON)}, pages 1--4, 2019.

\bibitem{walther2014classification}
Steffen Walther and Tim Guhl.
\newblock Classification of physical human-robot interaction scenarios to identify relevant requirements.
\newblock In {\em ISR/Robotik 2014; 41st International Symposium on Robotics}, pages 1--8. VDE, 2014.

\bibitem{muller2017subjective}
Sarah~L M{\"u}ller, Sebastian Stiehm, Sabina Jeschke, and Anja Richert.
\newblock Subjective stress in hybrid collaboration.
\newblock In {\em Social Robotics: 9th International Conference, ICSR 2017, Tsukuba, Japan, November 22-24, 2017, Proceedings 9}, pages 597--606. Springer, 2017.

\bibitem{8907351}
Uchenna~Emeoha Ogenyi, Jinguo Liu, Chenguang Yang, Zhaojie Ju, and Honghai Liu.
\newblock Physical human–robot collaboration: Robotic systems, learning methods, collaborative strategies, sensors, and actuators.
\newblock {\em IEEE Transactions on Cybernetics}, 51(4):1888--1901, 2021.

\bibitem{tsarouchi2016human}
Panagiota Tsarouchi, Sotiris Makris, and George Chryssolouris.
\newblock Human--robot interaction review and challenges on task planning and programming.
\newblock {\em International Journal of Computer Integrated Manufacturing}, 29(8):916--931, 2016.

\bibitem{zacharaki2020safety}
Angeliki Zacharaki, Ioannis Kostavelis, Antonios Gasteratos, and Ioannis Dokas.
\newblock Safety bounds in human robot interaction: A survey.
\newblock {\em Safety science}, 127:104667, 2020.

\bibitem{goodrich2008human}
Michael~A Goodrich, Alan~C Schultz, et~al.
\newblock Human--robot interaction: a survey.
\newblock {\em Foundations and Trends{\textregistered} in Human--Computer Interaction}, 1(3):203--275, 2008.

\bibitem{yan2014survey}
Haibin Yan, Marcelo~H Ang, and Aun~Neow Poo.
\newblock A survey on perception methods for human--robot interaction in social robots.
\newblock {\em International Journal of Social Robotics}, 6:85--119, 2014.

\bibitem{yang2018social}
Guang-Zhong Yang, Paolo Dario, and Danica Kragic.
\newblock Social robotics—trust, learning, and social interaction, 2018.

\bibitem{butepage2017human}
Judith B{\"u}tepage and Danica Kragic.
\newblock Human-robot collaboration: from psychology to social robotics.
\newblock {\em arXiv preprint arXiv:1705.10146}, 2017.

\bibitem{akalin2021reinforcement}
Neziha Akalin and Amy Loutfi.
\newblock Reinforcement learning approaches in social robotics.
\newblock {\em Sensors}, 21(4):1292, 2021.

\bibitem{costa2015using}
Sandra Costa, Hagen Lehmann, Kerstin Dautenhahn, Ben Robins, and Filomena Soares.
\newblock Using a humanoid robot to elicit body awareness and appropriate physical interaction in children with autism.
\newblock {\em International journal of social robotics}, 7:265--278, 2015.

\bibitem{1521743}
K.~Kosuge and Y.~Hirata.
\newblock Human-robot interaction.
\newblock In {\em 2004 IEEE International Conference on Robotics and Biomimetics}, pages 8--11, 2004.

\bibitem{pollmann2023entertainment}
Kathrin Pollmann, Wulf Loh, Nora Fronemann, and Daniel Ziegler.
\newblock Entertainment vs. manipulation: Personalized human-robot interaction between user experience and ethical design.
\newblock {\em Technological Forecasting and Social Change}, 189:122376, 2023.

\bibitem{6840111}
Steffen Walther and Tim Guhl.
\newblock Classification of physical human-robot interaction scenarios to identify relevant requirements.
\newblock In {\em ISR/Robotik 2014; 41st International Symposium on Robotics}, pages 1--8, 2014.

\bibitem{lasota2017survey}
Przemyslaw~A Lasota, Terrence Fong, Julie~A Shah, et~al.
\newblock A survey of methods for safe human-robot interaction.
\newblock {\em Foundations and Trends{\textregistered} in Robotics}, 5(4):261--349, 2017.

\bibitem{rahimi2018neural}
Hamed~N Rahimi, Ian Howard, and Lei Cui.
\newblock Neural impedance adaption for assistive human--robot interaction.
\newblock {\em Neurocomputing}, 290:50--59, 2018.

\bibitem{han2019admittance}
Yali Han, Songqing Zhu, Yiming Zhou, and Haitao Gao.
\newblock An admittance controller based on assistive torque estimation for a rehabilitation leg exoskeleton.
\newblock {\em Intelligent Service Robotics}, 12(4):381--391, 2019.

\bibitem{marban2019recurrent}
Arturo Marban, Vignesh Srinivasan, Wojciech Samek, Josep Fern{\'a}ndez, and Alicia Casals.
\newblock A recurrent convolutional neural network approach for sensorless force estimation in robotic surgery.
\newblock {\em Biomedical Signal Processing and Control}, 50:134--150, 2019.

\bibitem{9594838}
Jiuyun Xia and Kazuo Kiguchi.
\newblock Sensorless real-time force estimation in microsurgery robots using a time series convolutional neural network.
\newblock {\em IEEE Access}, 9:149447--149455, 2021.

\bibitem{kim2017impedance}
Yeoun~Jae Kim, Jong~Hyun Seo, Hong~Rae Kim, and Kwang~Gi Kim.
\newblock Impedance and admittance control for respiratory-motion compensation during robotic needle insertion--a preliminary test.
\newblock {\em The International Journal of Medical Robotics and Computer Assisted Surgery}, 13(4):e1795, 2017.

\bibitem{8333285}
Jason Fong and Mahdi Tavakoli.
\newblock Kinesthetic teaching of a therapist's behavior to a rehabilitation robot.
\newblock In {\em 2018 International Symposium on Medical Robotics (ISMR)}, pages 1--6, 2018.

\bibitem{mohammadi2020mixed}
Fatemeh Mohammadi~Amin, Maryam Rezayati, Hans~Wernher van~de Venn, and Hossein Karimpour.
\newblock A mixed-perception approach for safe human--robot collaboration in industrial automation.
\newblock {\em Sensors}, 20(21):6347, 2020.

\bibitem{app11093907}
Harsh Maithani, Juan~Antonio Corrales~Ramon, Laurent Lequievre, Youcef Mezouar, and Matthieu Alric.
\newblock Exoscarne: Assistive strategies for an industrial meat cutting system based on physical human-robot interaction.
\newblock {\em Applied Sciences}, 11(9), 2021.

\bibitem{yao2018sensorless}
Bitao Yao, Zude Zhou, Lihui Wang, Wenjun Xu, Quan Liu, and Aiming Liu.
\newblock Sensorless and adaptive admittance control of industrial robot in physical human- robot interaction.
\newblock {\em Robotics and Computer-Integrated Manufacturing}, 51:158--168, 2018.

\bibitem{10144527}
Christopher~Yee Wong, Lucas Vergez, and Wael Suleiman.
\newblock Vision-and tactile-based continuous multimodal intention and attention recognition for safer physical human–robot interaction.
\newblock {\em IEEE Transactions on Automation Science and Engineering}, pages 1--11, 2023.

\bibitem{8794065}
Sammy Christen, Stefan Stevšić, and Otmar Hilliges.
\newblock Demonstration-guided deep reinforcement learning of control policies for dexterous human-robot interaction.
\newblock In {\em 2019 International Conference on Robotics and Automation (ICRA)}, pages 2161--2167, 2019.

\bibitem{7858650}
Diego~Felipe Paez~Granados, Breno~A. Yamamoto, Hiroko Kamide, Jun Kinugawa, and Kazuhiro Kosuge.
\newblock Dance teaching by a robot: Combining cognitive and physical human–robot interaction for supporting the skill learning process.
\newblock {\em IEEE Robotics and Automation Letters}, 2(3):1452--1459, 2017.

\bibitem{akgun2012trajectories}
Baris Akgun, Maya Cakmak, Jae~Wook Yoo, and Andrea~Lockerd Thomaz.
\newblock Trajectories and keyframes for kinesthetic teaching: A human-robot interaction perspective.
\newblock In {\em Proceedings of the seventh annual ACM/IEEE international conference on Human-Robot Interaction}, pages 391--398, 2012.

\bibitem{topini2022variable}
Alberto Topini, William Sansom, Nicola Secciani, Lorenzo Bartalucci, Alessandro Ridolfi, and Benedetto Allotta.
\newblock Variable admittance control of a hand exoskeleton for virtual reality-based rehabilitation tasks.
\newblock {\em Frontiers in neurorobotics}, 15:188, 2022.

\bibitem{7759417}
Ali Ghadirzadeh, Judith Bütepage, Atsuto Maki, Danica Kragic, and Mårten Björkman.
\newblock A sensorimotor reinforcement learning framework for physical human-robot interaction.
\newblock In {\em 2016 IEEE/RSJ International Conference on Intelligent Robots and Systems (IROS)}, pages 2682--2688, 2016.

\bibitem{khoramshahi2019dynamical}
Mahdi Khoramshahi and Aude Billard.
\newblock A dynamical system approach to task-adaptation in physical human--robot interaction.
\newblock {\em Autonomous Robots}, 43:927--946, 2019.

\bibitem{7989334}
David Vogt, Simon Stepputtis, Steve Grehl, Bernhard Jung, and Heni Ben~Amor.
\newblock A system for learning continuous human-robot interactions from human-human demonstrations.
\newblock In {\em 2017 IEEE International Conference on Robotics and Automation (ICRA)}, pages 2882--2889, 2017.

\bibitem{7139393}
Marco Ewerton, Gerhard Neumann, Rudolf Lioutikov, Heni Ben~Amor, Jan Peters, and Guilherme Maeda.
\newblock Learning multiple collaborative tasks with a mixture of interaction primitives.
\newblock In {\em 2015 IEEE International Conference on Robotics and Automation (ICRA)}, pages 1535--1542, 2015.

\bibitem{tsiakas2017interactive}
Konstantinos Tsiakas, Michalis Papakostas, Michail Theofanidis, Morris Bell, Rada Mihalcea, Shouyi Wang, Mihai Burzo, and Fillia Makedon.
\newblock An interactive multisensing framework for personalized human robot collaboration and assistive training using reinforcement learning.
\newblock In {\em Proceedings of the 10th International Conference on PErvasive Technologies Related to Assistive Environments}, pages 423--427, 2017.

\bibitem{9340865}
Marina Kollmitz, Torsten Koller, Joschka Boedecker, and Wolfram Burgard.
\newblock Learning human-aware robot navigation from physical interaction via inverse reinforcement learning.
\newblock In {\em 2020 IEEE/RSJ International Conference on Intelligent Robots and Systems (IROS)}, pages 11025--11031, 2020.

\bibitem{9806047}
Mattia Leonori, Juan~M. Gandarias, and Arash Ajoudani.
\newblock Moca-s: A sensitive mobile collaborative robotic assistant exploiting low-cost capacitive tactile cover and whole-body control.
\newblock {\em IEEE Robotics and Automation Letters}, 7(3):7920--7927, 2022.

\bibitem{rozo2016learning}
Leonel Rozo, Joao Silverio, Sylvain Calinon, and Darwin~G Caldwell.
\newblock Learning controllers for reactive and proactive behaviors in human--robot collaboration.
\newblock {\em Frontiers in Robotics and AI}, 3:30, 2016.

\bibitem{9075439}
Wei He, Chengqian Xue, Xinbo Yu, Zhijun Li, and Chenguang Yang.
\newblock Admittance-based controller design for physical human–robot interaction in the constrained task space.
\newblock {\em IEEE Transactions on Automation Science and Engineering}, 17(4):1937--1949, 2020.

\bibitem{blancas2015effects}
Maria Blancas, Vasiliki Vouloutsi, Klaudia Grechuta, and Paul~FMJ Verschure.
\newblock Effects of the robot’s role on human-robot interaction in an educational scenario.
\newblock In {\em Biomimetic and Biohybrid Systems: 4th International Conference, Living Machines 2015, Barcelona, Spain, July 28-31, 2015, Proceedings 4}, pages 391--402. Springer, 2015.

\bibitem{henkemans2013using}
Olivier A~Blanson Henkemans, Bert~PB Bierman, Joris Janssen, Mark~A Neerincx, Rosemarijn Looije, Hanneke van~der Bosch, and Jeanine~AM van~der Giessen.
\newblock Using a robot to personalise health education for children with diabetes type 1: A pilot study.
\newblock {\em Patient education and counseling}, 92(2):174--181, 2013.

\bibitem{8625030}
Phuong~D.H. Nguyen, Fabrizio Bottarel, Ugo Pattacini, Matej Hoffmann, Lorenzo Natale, and Giorgio Metta.
\newblock Merging physical and social interaction for effective human-robot collaboration.
\newblock In {\em 2018 IEEE-RAS 18th International Conference on Humanoid Robots (Humanoids)}, pages 1--9, 2018.

\bibitem{robla2017working}
S.~Robla-Gómez, Victor~M. Becerra, J.~R. Llata, E.~González-Sarabia, C.~Torre-Ferrero, and J.~Pérez-Oria.
\newblock Working together: A review on safe human-robot collaboration in industrial environments.
\newblock {\em IEEE Access}, 5:26754--26773, 2017.

\bibitem{lima2019artificial}
Edirlei Soares~de Lima and Bruno Feij{\'o}.
\newblock Artificial intelligence in human-robot interaction.
\newblock In {\em Emotional Design in Human-Robot Interaction}, pages 187--199. Springer, 2019.

\bibitem{semeraro2023human}
Francesco Semeraro, Alexander Griffiths, and Angelo Cangelosi.
\newblock Human--robot collaboration and machine learning: A systematic review of recent research.
\newblock {\em Robotics and Computer-Integrated Manufacturing}, 79:102432, 2023.

\bibitem{vasic2013safety}
Milos Vasic and Aude Billard.
\newblock Safety issues in human-robot interactions.
\newblock In {\em 2013 ieee international conference on robotics and automation}, pages 197--204. IEEE, 2013.

\bibitem{papetti2022human}
Alessandra Papetti, Marianna Ciccarelli, Cecilia Scoccia, Giacomo Palmieri, and Michele Germani.
\newblock A human-oriented design process for collaborative robotics.
\newblock {\em International Journal of Computer Integrated Manufacturing}, pages 1--23, 2022.

\bibitem{boschetti2022human}
Giovanni Boschetti, Maurizio Faccio, and Irene Granata.
\newblock Human-centered design for productivity and safety in collaborative robots cells: A new methodological approach.
\newblock {\em Electronics}, 12(1):167, 2022.

\bibitem{gualtieri2022development}
Luca Gualtieri, Erwin Rauch, and Renato Vidoni.
\newblock Development and validation of guidelines for safety in human-robot collaborative assembly systems.
\newblock {\em Computers \& Industrial Engineering}, 163:107801, 2022.

\bibitem{maurice2017human}
Pauline Maurice, Vincent Padois, Yvan Measson, and Philippe Bidaud.
\newblock Human-oriented design of collaborative robots.
\newblock {\em International Journal of Industrial Ergonomics}, 57:88--102, 2017.

\bibitem{10000222}
Carlotta Sartore, Lorenzo Rapetti, and Daniele Pucci.
\newblock Optimization of humanoid robot designs for human-robot ergonomic payload lifting.
\newblock In {\em 2022 IEEE-RAS 21st International Conference on Humanoid Robots (Humanoids)}, pages 722--729, 2022.

\bibitem{sosa2018robot}
Ricardo Sosa, Miguel Montiel, Eduardo~B Sandoval, Rajesh~E Mohan, et~al.
\newblock Robot ergonomics: Towards human-centred and robot-inclusive design.
\newblock In {\em DS 92: Proceedings of the DESIGN 2018 15th International Design Conference}, pages 2323--2334, 2018.

\bibitem{gualtieri2020safety}
Luca Gualtieri, Erwin Rauch, Renato Vidoni, and Dominik~T Matt.
\newblock Safety, ergonomics and efficiency in human-robot collaborative assembly: design guidelines and requirements.
\newblock {\em Procedia CIRP}, 91:367--372, 2020.

\bibitem{rubagotti2022perceived}
Matteo Rubagotti, Inara Tusseyeva, Sara Baltabayeva, Danna Summers, and Anara Sandygulova.
\newblock Perceived safety in physical human--robot interaction—a survey.
\newblock {\em Robotics and Autonomous Systems}, 151:104047, 2022.

\bibitem{akalin2022you}
Neziha Akalin, Annica Kristoffersson, and Amy Loutfi.
\newblock Do you feel safe with your robot? factors influencing perceived safety in human-robot interaction based on subjective and objective measures.
\newblock {\em International journal of human-computer studies}, 158:102744, 2022.

\bibitem{4581481}
Olesya Ogorodnikova.
\newblock Methodology of safety for a human robot interaction designing stage.
\newblock In {\em 2008 Conference on Human System Interactions}, pages 452--457, 2008.

\bibitem{lim2021social}
Velvetina Lim, Maki Rooksby, and Emily~S Cross.
\newblock Social robots on a global stage: establishing a role for culture during human--robot interaction.
\newblock {\em International Journal of Social Robotics}, 13(6):1307--1333, 2021.

\bibitem{alzahrani2022exploring}
Abdullah Alzahrani, Simon Robinson, and Muneeb Ahmad.
\newblock Exploring factors affecting user trust across different human-robot interaction settings and cultures.
\newblock In {\em Proceedings of the 10th International Conference on Human-Agent Interaction}, pages 123--131, 2022.

\bibitem{lu2022mental}
Lu~Lu, Ziyang Xie, Hanwen Wang, Li~Li, and Xu~Xu.
\newblock Mental stress and safety awareness during human-robot collaboration-review.
\newblock {\em Applied Ergonomics}, 105:103832, 2022.

\bibitem{pervez2008safe}
Aslam Pervez and Jeha Ryu.
\newblock Safe physical human robot interaction-past, present and future.
\newblock {\em Journal of Mechanical Science and Technology}, 22:469--483, 2008.

\bibitem{she2016design}
Yu~She, Hai-Jun Su, Cheng Lai, and Deshan Meng.
\newblock Design and prototype of a tunable stiffness arm for safe human-robot interaction.
\newblock In {\em International design engineering technical conferences and computers and information in engineering conference}, volume 50169, page V05BT07A063. American Society of Mechanical Engineers, 2016.

\bibitem{van2009compliant}
Ronald Van~Ham, Thomas~G Sugar, Bram Vanderborght, Kevin~W Hollander, and Dirk Lefeber.
\newblock Compliant actuator designs.
\newblock {\em IEEE Robotics \& Automation Magazine}, 16(3):81--94, 2009.

\bibitem{zinn2004playing}
Michael Zinn, Oussama Khatib, Bernard Roth, and J~Kenneth Salisbury.
\newblock Playing it safe [human-friendly robots].
\newblock {\em IEEE Robotics \& Automation Magazine}, 11(2):12--21, 2004.

\bibitem{pratt1995series}
Gill~A Pratt and Matthew~M Williamson.
\newblock Series elastic actuators.
\newblock In {\em Proceedings 1995 IEEE/RSJ International Conference on Intelligent Robots and Systems. Human Robot Interaction and Cooperative Robots}, volume~1, pages 399--406. IEEE, 1995.

\bibitem{1570172}
G.~Tonietti, R.~Schiavi, and A.~Bicchi.
\newblock Design and control of a variable stiffness actuator for safe and fast physical human/robot interaction.
\newblock In {\em Proceedings of the 2005 IEEE International Conference on Robotics and Automation}, pages 526--531, 2005.

\bibitem{bicchi2005variable}
Antonio Bicchi, Giovanni Tonietti, Michele Bavaro, and Marco Piccigallo.
\newblock Variable stiffness actuators for fast and safe motion control.
\newblock In {\em Robotics Research. The Eleventh International Symposium: With 303 Figures}, pages 527--536. Springer, 2005.

\bibitem{8794236}
David~V. Gealy, Stephen McKinley, Brent Yi, Philipp Wu, Phillip~R. Downey, Greg Balke, Allan Zhao, Menglong Guo, Rachel Thomasson, Anthony Sinclair, Peter Cuellar, Zoe McCarthy, and Pieter Abbeel.
\newblock Quasi-direct drive for low-cost compliant robotic manipulation.
\newblock In {\em 2019 International Conference on Robotics and Automation (ICRA)}, pages 437--443, 2019.

\bibitem{7403902}
Gavin Kenneally, Avik De, and D.~E. Koditschek.
\newblock Design principles for a family of direct-drive legged robots.
\newblock {\em IEEE Robotics and Automation Letters}, 1(2):900--907, 2016.

\bibitem{525724}
K.~Suita, Y.~Yamada, N.~Tsuchida, K.~Imai, H.~Ikeda, and N.~Sugimoto.
\newblock A failure-to-safety "kyozon" system with simple contact detection and stop capabilities for safe human-autonomous robot coexistence.
\newblock In {\em Proceedings of 1995 IEEE International Conference on Robotics and Automation}, volume~3, pages 3089--3096 vol.3, 1995.

\bibitem{lim1999collision}
Hun-Ok Lim and Kazuo Tanie.
\newblock Collision-tolerant control of human-friendly robot with viscoelastic trunk.
\newblock {\em IEEE/ASME transactions on mechatronics}, 4(4):417--427, 1999.

\bibitem{7353705}
Joohyung Kim, Alexander Alspach, and Katsu Yamane.
\newblock 3d printed soft skin for safe human-robot interaction.
\newblock In {\em 2015 IEEE/RSJ International Conference on Intelligent Robots and Systems (IROS)}, pages 2419--2425, 2015.

\bibitem{chang2015interaction}
Wan-Ling Chang and Selma {\v{S}}abanovi{\'c}.
\newblock Interaction expands function: Social shaping of the therapeutic robot paro in a nursing home.
\newblock In {\em Proceedings of the Tenth Annual ACM/IEEE International Conference on Human-Robot Interaction}, pages 343--350, 2015.

\bibitem{6907362}
Ronghuai Qi, Tin~Lun Lam, and Yangsheng Xu.
\newblock Mechanical design and implementation of a soft inflatable robot arm for safe human-robot interaction.
\newblock In {\em 2014 IEEE International Conference on Robotics and Automation (ICRA)}, pages 3490--3495, 2014.

\bibitem{bicchi2004fast}
Antonio Bicchi and Giovanni Tonietti.
\newblock Fast and" soft-arm" tactics [robot arm design].
\newblock {\em IEEE Robotics \& Automation Magazine}, 11(2):22--33, 2004.

\bibitem{5756872}
Rainer Bischoff, Johannes Kurth, Guenter Schreiber, Ralf Koeppe, Alin Albu-Schaeffer, Alexander Beyer, Oliver Eiberger, Sami Haddadin, Andreas Stemmer, Gerhard Grunwald, and Gerhard Hirzinger.
\newblock The kuka-dlr lightweight robot arm - a new reference platform for robotics research and manufacturing.
\newblock In {\em ISR 2010 (41st International Symposium on Robotics) and ROBOTIK 2010 (6th German Conference on Robotics)}, pages 1--8, 2010.

\bibitem{de2006collision}
Alessandro De~Luca, Alin Albu-Schaffer, Sami Haddadin, and Gerd Hirzinger.
\newblock Collision detection and safe reaction with the dlr-iii lightweight manipulator arm.
\newblock In {\em 2006 IEEE/RSJ International Conference on Intelligent Robots and Systems}, pages 1623--1630. IEEE, 2006.

\bibitem{rybski2012sensor}
Paul Rybski, Peter Anderson-Sprecher, Daniel Huber, Chris Niessl, and Reid Simmons.
\newblock Sensor fusion for human safety in industrial workcells.
\newblock In {\em 2012 IEEE/RSJ International Conference on Intelligent Robots and Systems}, pages 3612--3619. IEEE, 2012.

\bibitem{fritzsche2011tactile}
Markus Fritzsche, Norbert Elkmann, and Erik Schulenburg.
\newblock Tactile sensing: A key technology for safe physical human robot interaction.
\newblock In {\em Proceedings of the 6th International Conference on Human-robot Interaction}, pages 139--140, 2011.

\bibitem{8793258}
Emmanuel Dean-Leon, J.~Rogelio Guadarrama-Olvera, Florian Bergner, and Gordon Cheng.
\newblock Whole-body active compliance control for humanoid robots with robot skin.
\newblock In {\em 2019 International Conference on Robotics and Automation (ICRA)}, pages 5404--5410, 2019.

\bibitem{maiolino2013.iCubSkin}
Perla Maiolino, Marco Maggiali, Giorgio Cannata, Giorgio Metta, and Lorenzo Natale.
\newblock A flexible and robust large scale capacitive tactile system for robots.
\newblock {\em IEEE Sensors Journal}, 13(10):3910--3917, 2013.

\bibitem{9197365}
Isabella Huang and Ruzena Bajcsy.
\newblock High resolution soft tactile interface for physical human-robot interaction.
\newblock In {\em 2020 IEEE International Conference on Robotics and Automation (ICRA)}, pages 1705--1711, 2020.

\bibitem{7346450}
A.~Cirillo, F.~Ficuciello, C.~Natale, S.~Pirozzi, and L.~Villani.
\newblock A conformable force/tactile skin for physical human–robot interaction.
\newblock {\em IEEE Robotics and Automation Letters}, 1(1):41--48, 2016.

\bibitem{haddadin2012making}
Sami Haddadin, Simon Haddadin, Augusto Khoury, Tim Rokahr, Sven Parusel, Rainer Burgkart, Antonio Bicchi, and Alin Albu-Sch{\"a}ffer.
\newblock On making robots understand safety: Embedding injury knowledge into control.
\newblock {\em The International Journal of Robotics Research}, 31(13):1578--1602, 2012.

\bibitem{morato2013safe}
Carlos Morato, Krishnanand Kaipa, Boxuan Zhao, and Satyandra~K Gupta.
\newblock Safe human robot interaction by using exteroceptive sensing based human modeling.
\newblock In {\em International Design Engineering Technical Conferences and Computers and Information in Engineering Conference}, volume 55850, page V02AT02A073. American Society of Mechanical Engineers, 2013.

\bibitem{kulic2007pre}
Dana Kuli{\'c} and Elizabeth Croft.
\newblock Pre-collision safety strategies for human-robot interaction.
\newblock {\em Autonomous Robots}, 22:149--164, 2007.

\bibitem{mainprice2013human}
Jim Mainprice and Dmitry Berenson.
\newblock Human-robot collaborative manipulation planning using early prediction of human motion.
\newblock In {\em 2013 IEEE/RSJ International Conference on Intelligent Robots and Systems}, pages 299--306. IEEE, 2013.

\bibitem{5980248}
Hao Ding, Gunther Reißig, Kurniawan Wijaya, Dino Bortot, Klaus Bengler, and Olaf Stursberg.
\newblock Human arm motion modeling and long-term prediction for safe and efficient human-robot-interaction.
\newblock In {\em 2011 IEEE International Conference on Robotics and Automation}, pages 5875--5880, 2011.

\bibitem{9300047}
Qinghua Li, Zhao Zhang, Yue You, Yaqi Mu, and Chao Feng.
\newblock Data driven models for human motion prediction in human-robot collaboration.
\newblock {\em IEEE Access}, 8:227690--227702, 2020.

\bibitem{CHOI2022102258}
Sung~Ho Choi, Kyeong-Beom Park, Dong~Hyeon Roh, Jae~Yeol Lee, Mustafa Mohammed, Yalda Ghasemi, and Heejin Jeong.
\newblock An integrated mixed reality system for safety-aware human-robot collaboration using deep learning and digital twin generation.
\newblock {\em Robotics and Computer-Integrated Manufacturing}, 73:102258, 2022.

\bibitem{admoni2017social}
Henny Admoni and Brian Scassellati.
\newblock Social eye gaze in human-robot interaction: a review.
\newblock {\em Journal of Human-Robot Interaction}, 6(1):25--63, 2017.

\bibitem{8593580}
Akanksha Saran, Srinjoy Majumdar, Elaine~Schaertl Short, Andrea Thomaz, and Scott Niekum.
\newblock Human gaze following for human-robot interaction.
\newblock In {\em 2018 IEEE/RSJ International Conference on Intelligent Robots and Systems (IROS)}, pages 8615--8621, 2018.

\bibitem{upasani2023eye}
Satyajit Upasani, Divya Srinivasan, Qi~Zhu, Jing Du, and Alexander Leonessa.
\newblock Eye-tracking in physical human--robot interaction: Mental workload and performance prediction.
\newblock {\em Human factors}, page 00187208231204704, 2023.

\bibitem{haji2018exploiting}
Alireza Haji~Fathaliyan, Xiaoyu Wang, and Veronica~J Santos.
\newblock Exploiting three-dimensional gaze tracking for action recognition during bimanual manipulation to enhance human--robot collaboration.
\newblock {\em Frontiers in Robotics and AI}, 5:25, 2018.

\bibitem{8793657}
Eleonora Mariotti, Emanuele Magrini, and Alessandro~De Luca.
\newblock Admittance control for human-robot interaction using an industrial robot equipped with a f/t sensor.
\newblock In {\em 2019 International Conference on Robotics and Automation (ICRA)}, pages 6130--6136, 2019.

\bibitem{4650764}
Sami Haddadin, Alin Albu-Schaffer, Alessandro De~Luca, and Gerd Hirzinger.
\newblock Collision detection and reaction: A contribution to safe physical human-robot interaction.
\newblock In {\em 2008 IEEE/RSJ International Conference on Intelligent Robots and Systems}, pages 3356--3363, 2008.

\bibitem{li2021nonlinear}
Yi~Li, Yanhui Li, Mingchao Zhu, Zhenbang Xu, and Deqiang Mu.
\newblock A nonlinear momentum observer for sensorless robot collision detection under model uncertainties.
\newblock {\em Mechatronics}, 78:102603, 2021.

\bibitem{6899348}
Przemyslaw~A. Lasota, Gregory~F. Rossano, and Julie~A. Shah.
\newblock Toward safe close-proximity human-robot interaction with standard industrial robots.
\newblock In {\em 2014 IEEE International Conference on Automation Science and Engineering (CASE)}, pages 339--344, 2014.

\bibitem{lasota2015analyzing}
Przemyslaw~A Lasota and Julie~A Shah.
\newblock Analyzing the effects of human-aware motion planning on close-proximity human--robot collaboration.
\newblock {\em Human factors}, 57(1):21--33, 2015.

\bibitem{6197743}
Emrah~Akin Sisbot and Rachid Alami.
\newblock A human-aware manipulation planner.
\newblock {\em IEEE Transactions on Robotics}, 28(5):1045--1057, 2012.

\bibitem{7487584}
Rafi Hayne, Ruikun Luo, and Dmitry Berenson.
\newblock Considering avoidance and consistency in motion planning for human-robot manipulation in a shared workspace.
\newblock In {\em 2016 IEEE International Conference on Robotics and Automation (ICRA)}, pages 3948--3954, 2016.

\bibitem{9913366}
Marco Faroni, Manuel Beschi, and Nicola Pedrocchi.
\newblock Safety-aware time-optimal motion planning with uncertain human state estimation.
\newblock {\em IEEE Robotics and Automation Letters}, 7(4):12219--12226, 2022.

\bibitem{de2007reactive}
Agostino De~Santis, Bruno Siciliano, et~al.
\newblock Reactive collision avoidance for safer human--robot interaction.
\newblock In {\em 5th IARP/IEEE RAS/EURON workshop on technical challenges for dependable robots in human environments}, volume~1. Citeseer, 2007.

\bibitem{8004480}
Yiwei Wang, Yixuan Sheng, Ji~Wang, and Wenlong Zhang.
\newblock Optimal collision-free robot trajectory generation based on time series prediction of human motion.
\newblock {\em IEEE Robotics and Automation Letters}, 3(1):226--233, 2018.

\bibitem{haddadin2013almost}
Sami Haddadin, Sven Parusel, Rico Belder, and Alin Albu-Sch{\"a}ffer.
\newblock It is (almost) all about human safety: A novel paradigm for robot design, control, and planning.
\newblock In {\em Computer Safety, Reliability, and Security: 32nd International Conference, SAFECOMP 2013, Toulouse, France, September 24-27, 2013. Proceedings 32}, pages 202--215. Springer, 2013.

\bibitem{haddadin2011dynamic}
Sami Haddadin, Rico Belder, and Alin Albu-Sch{\"a}ffer.
\newblock Dynamic motion planning for robots in partially unknown environments.
\newblock {\em IFAC Proceedings Volumes}, 44(1):6842--6850, 2011.

\bibitem{9488306}
J.~Micah Prendergast, Stephan Balvert, Tom Driessen, Ajay Seth, and Luka Peternel.
\newblock Biomechanics aware collaborative robot system for delivery of safe physical therapy in shoulder rehabilitation.
\newblock {\em IEEE Robotics and Automation Letters}, 6(4):7177--7184, 2021.

\bibitem{9034996}
Milad Shafiee, Giulio Romualdi, Stefano Dafarra, Francisco Javier~Andrade Chavez, and Daniele Pucci.
\newblock Online dcm trajectory generation for push recovery of torque-controlled humanoid robots.
\newblock In {\em 2019 IEEE-RAS 19th International Conference on Humanoid Robots (Humanoids)}, pages 671--678, 2019.

\bibitem{7079531}
Andrea~Maria Zanchettin, Nicola~Maria Ceriani, Paolo Rocco, Hao Ding, and Björn Matthias.
\newblock Safety in human-robot collaborative manufacturing environments: Metrics and control.
\newblock {\em IEEE Transactions on Automation Science and Engineering}, 13(2):882--893, 2016.

\bibitem{oleinikov2021safety}
Artemiy Oleinikov, Sanzhar Kusdavletov, Almas Shintemirov, and Matteo Rubagotti.
\newblock Safety-aware nonlinear model predictive control for physical human-robot interaction.
\newblock {\em IEEE Robotics and Automation Letters}, 6(3):5665--5672, 2021.

\bibitem{vick2013safe}
Axel Vick, Dragoljub Surdilovic, and J{\"o}rg Kr{\"u}ger.
\newblock Safe physical human-robot interaction with industrial dual-arm robots.
\newblock In {\em 9th International Workshop on Robot Motion and Control}, pages 264--269. IEEE, 2013.

\bibitem{bian2018improving}
Feifei Bian, Danmei Ren, Ruifeng Li, and Peidong Liang.
\newblock Improving stability in physical human--robot interaction by estimating human hand stiffness and a vibration index.
\newblock {\em Industrial Robot: the international journal of robotics research and application}, 2018.

\bibitem{silvera2015artificial}
David Silvera-Tawil, David Rye, and Mari Velonaki.
\newblock Artificial skin and tactile sensing for socially interactive robots: A review.
\newblock {\em Robotics and Autonomous Systems}, 63:230--243, 2015.

\bibitem{li2022multifunctional}
Guozhen Li, Shiqiang Liu, Qian Mao, and Rong Zhu.
\newblock Multifunctional electronic skins enable robots to safely and dexterously interact with human.
\newblock {\em Advanced Science}, 9(11):2104969, 2022.

\bibitem{6631284}
Federica Ferraguti, Cristian Secchi, and Cesare Fantuzzi.
\newblock A tank-based approach to impedance control with variable stiffness.
\newblock In {\em 2013 IEEE International Conference on Robotics and Automation}, pages 4948--4953, 2013.

\bibitem{cortez2021safe}
Wenceslao~Shaw Cortez, Christos~K Verginis, and Dimos~V Dimarogonas.
\newblock Safe, passive control for mechanical systems with application to physical human-robot interactions.
\newblock In {\em 2021 IEEE International Conference on Robotics and Automation (ICRA)}, pages 3836--3842. IEEE, 2021.

\bibitem{chen2022human}
Jingdong Chen and Paul~I Ro.
\newblock Human intention-oriented variable admittance control with power envelope regulation in physical human-robot interaction.
\newblock {\em Mechatronics}, 84:102802, 2022.

\bibitem{8685113}
Gitae Kang, Hyun~Seok Oh, Joon~Kyue Seo, Uikyum Kim, and Hyouk~Ryeol Choi.
\newblock Variable admittance control of robot manipulators based on human intention.
\newblock {\em IEEE/ASME Transactions on Mechatronics}, 24(3):1023--1032, 2019.

\bibitem{Zhou2014Stabilizer}
Chengxu Zhou, Zhibin Li, Juan Castano, Houman Dallali, Nikos~G. Tsagarakis, and Darwin~G. Caldwell.
\newblock A passivity based compliance stabilizer for humanoid robots.
\newblock In {\em 2014 IEEE International Conference on Robotics and Automation (ICRA)}, pages 1487--1492, 2014.

\bibitem{tirupachuri2020towards}
Yeshasvi Tirupachuri, Gabriele Nava, Claudia Latella, Diego Ferigo, Lorenzo Rapetti, Luca Tagliapietra, Francesco Nori, and Daniele Pucci.
\newblock Towards partner-aware humanoid robot control under physical interactions.
\newblock In {\em Intelligent Systems and Applications: Proceedings of the 2019 Intelligent Systems Conference (IntelliSys) Volume 2}, pages 1073--1092. Springer, 2020.

\bibitem{7139727}
Michael Shomin, Jodi Forlizzi, and Ralph Hollis.
\newblock Sit-to-stand assistance with a balancing mobile robot.
\newblock In {\em 2015 IEEE International Conference on Robotics and Automation (ICRA)}, pages 3795--3800, 2015.

\bibitem{8968546}
Zhongyu Li and Ralph Hollis.
\newblock Toward a ballbot for physically leading people: A human-centered approach.
\newblock In {\em 2019 IEEE/RSJ International Conference on Intelligent Robots and Systems (IROS)}, pages 4827--4833, 2019.

\bibitem{kobayashi2022whole}
Taisuke Kobayashi, Emmanuel Dean-Leon, Julio~Rogelio Guadarrama-Olvera, Florian Bergner, and Gordon Cheng.
\newblock Whole-body multicontact haptic human--humanoid interaction based on leader--follower switching: A robot dance of the “box step”.
\newblock {\em Advanced Intelligent Systems}, 4(2):2100038, 2022.

\bibitem{englsberger2015DCM}
Johannes Englsberger, Christian Ott, and Alin Albu-Schäffer.
\newblock Three-dimensional bipedal walking control based on divergent component of motion.
\newblock {\em IEEE Transactions on Robotics}, 31(2):355--368, 2015.

\bibitem{1243931}
J.L. Drury, J.~Scholtz, and H.A. Yanco.
\newblock Awareness in human-robot interactions.
\newblock In {\em SMC'03 Conference Proceedings. 2003 IEEE International Conference on Systems, Man and Cybernetics. Conference Theme - System Security and Assurance (Cat. No.03CH37483)}, volume~1, pages 912--918 vol.1, 2003.

\bibitem{russell2016artificial}
S.~Russell and P.~Norvig.
\newblock {\em Artificial Intelligence: A Modern Approach}.
\newblock Always learning. Pearson, 2016.

\bibitem{durrant2012integration}
Hugh~F Durrant-Whyte.
\newblock {\em Integration, coordination and control of multi-sensor robot systems}, volume~36.
\newblock Springer Science \& Business Media, 2012.

\bibitem{Haddadin2016.pHRI}
Sami Haddadin and Elizabeth Croft.
\newblock {\em Physical Human--Robot Interaction}, pages 1835--1874.
\newblock Springer International Publishing, Cham, 2016.

\bibitem{Grunwald2003.TorqueControl}
G.~Grunwald, G.~Schreiber, A.~Albu-Schaffer, and G.~Hirzinger.
\newblock Programming by touch: the different way of human-robot interaction.
\newblock {\em IEEE Transactions on Industrial Electronics}, 50(4):659--666, 2003.

\bibitem{5152664}
Vincent Duchaine and Clement Gosselin.
\newblock Safe, stable and intuitive control for physical human-robot interaction.
\newblock In {\em 2009 IEEE International Conference on Robotics and Automation}, pages 3383--3388, 2009.

\bibitem{8328912}
Zhijun Li, Bo~Huang, Zhifeng Ye, Mingdi Deng, and Chenguang Yang.
\newblock Physical human–robot interaction of a robotic exoskeleton by admittance control.
\newblock {\em IEEE Transactions on Industrial Electronics}, 65(12):9614--9624, 2018.

\bibitem{7989338}
Chiara~Talignani Landi, Federica Ferraguti, Lorenzo Sabattini, Cristian Secchi, and Cesare Fantuzzi.
\newblock Admittance control parameter adaptation for physical human-robot interaction.
\newblock In {\em 2017 IEEE International Conference on Robotics and Automation (ICRA)}, pages 2911--2916, 2017.

\bibitem{8307450}
Hsieh-Yu Li, Ishara Paranawithana, Liangjing Yang, Terence Sey~Kiat Lim, Shaohui Foong, Foo~Cheong Ng, and U-Xuan Tan.
\newblock Stable and compliant motion of physical human–robot interaction coupled with a moving environment using variable admittance and adaptive control.
\newblock {\em IEEE Robotics and Automation Letters}, 3(3):2493--2500, 2018.

\bibitem{Wong2022.TouchSemantics}
Christopher~Yee Wong, Saeid Samadi, Wael Suleiman, Abderrahmane Kheddar, and Christopher~Yee Wong.
\newblock Touch semantics for intuitive physical manipulation of humanoids.
\newblock {\em IEEE transactions on human-machine systems.}, 52(6), 2022-12.

\bibitem{Holgado2019}
Alexis~C. Holgado, Nicola Piga, Tito~Pradhono Tomo, Giulia Vezzani, Alexander Schmitz, Lorenzo Natale, and Shigeki Sugano.
\newblock Magnetic 3-axis soft and sensitive fingertip sensors integration for the icub humanoid robot.
\newblock In {\em 2019 IEEE-RAS 19th International Conference on Humanoid Robots (Humanoids)}, pages 1--8, 2019.

\bibitem{Holgado2020}
Alexis~Carlos Holgado, Tito~Pradhono Tomo, Sophon Somlor, and Shigeki Sugano.
\newblock A multimodal, adjustable sensitivity, digital 3-axis skin sensor module.
\newblock {\em Sensors}, 20(11), 2020.

\bibitem{albini2020pressure}
Alessandro Albini and Giorgio Cannata.
\newblock Pressure distribution classification and segmentation of human hands in contact with the robot body.
\newblock {\em The International Journal of Robotics Research}, 39(6):668--687, 2020.

\bibitem{Leonori2022}
Mattia Leonori, Juan~M. Gandarias, and Arash Ajoudani.
\newblock Moca-s: A sensitive mobile collaborative robotic assistant exploiting low-cost capacitive tactile cover and whole-body control.
\newblock {\em IEEE Robotics and Automation Letters}, 7(3):7920--7927, 2022.

\bibitem{GordonCheng.skin}
P.~Mittendorfer, E.~Yoshida, and G.~Cheng.
\newblock Realizing whole-body tactile interactions with a self-organizing, multi-modal artificial skin on a humanoid robot.
\newblock {\em Advanced Robotics}, 29(1):51--67, 2015.

\bibitem{armleder2022interactive}
Simon Armleder, Emmanuel Dean-Leon, Florian Bergner, and Gordon Cheng.
\newblock Interactive force control based on multimodal robot skin for physical human- robot collaboration.
\newblock {\em Advanced Intelligent Systems}, 4(2):2100047, 2022.

\bibitem{5152595}
Vincent Duchaine, Nicolas Lauzier, Mathieu Baril, Marc-Antoine Lacasse, and Clement Gosselin.
\newblock A flexible robot skin for safe physical human robot interaction.
\newblock In {\em 2009 IEEE International Conference on Robotics and Automation}, pages 3676--3681, 2009.

\bibitem{Teyssier2021}
Marc Teyssier, Brice Parilusyan, Anne Roudaut, and Jürgen Steimle.
\newblock Human-like artificial skin sensor for physical human-robot interaction.
\newblock In {\em 2021 IEEE International Conference on Robotics and Automation (ICRA)}, pages 3626--3633, 2021.

\bibitem{wei2022flexible}
Shan Wei, Yijian Liu, Lina Yang, Haicheng Wang, Haoran Niu, Chao Zhou, Yanyan Wang, Qiuquan Guo, and Da~Chen.
\newblock Flexible large e-skin array based on patterned laser-induced graphene for tactile perception.
\newblock {\em Sensors and Actuators A: Physical}, 334:113308, 2022.

\bibitem{9094690}
Teng Xue, Weiming Wang, Jin Ma, Wenhai Liu, Zhenyu Pan, and Mingshuo Han.
\newblock Progress and prospects of multimodal fusion methods in physical human–robot interaction: A review.
\newblock {\em IEEE Sensors Journal}, 20(18):10355--10370, 2020.

\bibitem{Magrini2014.ContactEstimation}
Emanuele Magrini, Fabrizio Flacco, and Alessandro De~Luca.
\newblock Estimation of contact forces using a virtual force sensor.
\newblock In {\em 2014 IEEE/RSJ International Conference on Intelligent Robots and Systems}, pages 2126--2133, 2014.

\bibitem{wahrburg2015cartesian}
Arne Wahrburg, Bj{\"o}rn Matthias, and Hao Ding.
\newblock Cartesian contact force estimation for robotic manipulators-a fault isolation perspective.
\newblock {\em IFAC-PapersOnLine}, 48(21):1232--1237, 2015.

\bibitem{Zhang2020.vision-based-tactile-sensing}
Yazhan Zhang, Guanlan Zhang, Yipai Du, and Michael Yu~Wang.
\newblock Vtacarm. a vision-based tactile sensing augmented robotic arm with application to human-robot interaction.
\newblock In {\em 2020 IEEE 16th International Conference on Automation Science and Engineering (CASE)}, pages 35--42, 2020.

\bibitem{su2021deep}
Hang Su, Wen Qi, Zhijun Li, Ziyang Chen, Giancarlo Ferrigno, and Elena De~Momi.
\newblock Deep neural network approach in emg-based force estimation for human--robot interaction.
\newblock {\em IEEE Transactions on Artificial Intelligence}, 2(5):404--412, 2021.

\bibitem{7844516}
Stavros Grafakos, Fotios Dimeas, and Nikos Aspragathos.
\newblock Variable admittance control in phri using emg-based arm muscles co-activation.
\newblock In {\em 2016 IEEE International Conference on Systems, Man, and Cybernetics (SMC)}, pages 001900--001905, 2016.

\bibitem{bandara2018noninvasive}
DSV Bandara, Jumpei Arata, and Kazuo Kiguchi.
\newblock A noninvasive brain--computer interface approach for predicting motion intention of activities of daily living tasks for an upper-limb wearable robot.
\newblock {\em International Journal of Advanced Robotic Systems}, 15(2):1729881418767310, 2018.

\bibitem{roda2021comparison}
Luis Roda-Sanchez, Celia Garrido-Hidalgo, Arturo~S Garc{\'\i}a, Teresa Olivares, and Antonio Fern{\'a}ndez-Caballero.
\newblock Comparison of rgb-d and imu-based gesture recognition for human-robot interaction in remanufacturing.
\newblock {\em The International Journal of Advanced Manufacturing Technology}, pages 1--13, 2021.

\bibitem{campbell2020learning}
Joseph Campbell and Katsu Yamane.
\newblock Learning whole-body human-robot haptic interaction in social contexts.
\newblock In {\em 2020 IEEE International Conference on Robotics and Automation (ICRA)}, pages 10177--10183. IEEE, 2020.

\bibitem{bingol2020performing}
Mustafa~Can Bingol and Omur Aydogmus.
\newblock Performing predefined tasks using the human--robot interaction on speech recognition for an industrial robot.
\newblock {\em Engineering Applications of Artificial Intelligence}, 95:103903, 2020.

\bibitem{ashok2022collaborative}
K~Ashok, Mohd Ashraf, J~Thimmia~Raja, Md~Zair Hussain, Dinesh~Kumar Singh, and Anandakumar Haldorai.
\newblock Collaborative analysis of audio-visual speech synthesis with sensor measurements for regulating human--robot interaction.
\newblock {\em International Journal of System Assurance Engineering and Management}, pages 1--8, 2022.

\bibitem{gao2022tactile}
Shuo Gao, Yanning Dai, and Arokia Nathan.
\newblock Tactile and vision perception for intelligent humanoids.
\newblock {\em Advanced Intelligent Systems}, 4(2):2100074, 2022.

\bibitem{7139504}
Emanuele Magrini, Fabrizio Flacco, and Alessandro De~Luca.
\newblock Control of generalized contact motion and force in physical human-robot interaction.
\newblock In {\em 2015 IEEE International Conference on Robotics and Automation (ICRA)}, pages 2298--2304, 2015.

\bibitem{agravante2014collaborative}
Don~Joven Agravante, Andrea Cherubini, Antoine Bussy, Pierre Gergondet, and Abderrahmane Kheddar.
\newblock Collaborative human-humanoid carrying using vision and haptic sensing.
\newblock In {\em 2014 IEEE international conference on robotics and automation (ICRA)}, pages 607--612. IEEE, 2014.

\bibitem{6906917}
Don~Joven Agravante, Andrea Cherubini, Antoine Bussy, Pierre Gergondet, and Abderrahmane Kheddar.
\newblock Collaborative human-humanoid carrying using vision and haptic sensing.
\newblock In {\em 2014 IEEE International Conference on Robotics and Automation (ICRA)}, pages 607--612, 2014.

\bibitem{1244649}
H.~Kawamoto, Suwoong Lee, S.~Kanbe, and Y.~Sankai.
\newblock Power assist method for hal-3 using emg-based feedback controller.
\newblock In {\em SMC'03 Conference Proceedings. 2003 IEEE International Conference on Systems, Man and Cybernetics. Conference Theme - System Security and Assurance (Cat. No.03CH37483)}, volume~2, pages 1648--1653 vol.2, 2003.

\bibitem{7926461}
Kai Gui, Honghai Liu, and Dingguo Zhang.
\newblock Toward multimodal human–robot interaction to enhance active participation of users in gait rehabilitation.
\newblock {\em IEEE Transactions on Neural Systems and Rehabilitation Engineering}, 25(11):2054--2066, 2017.

\bibitem{al2021improving}
Ali Al-Yacoub, YC~Zhao, William Eaton, Yee~Mey Goh, and Niels Lohse.
\newblock Improving human robot collaboration through force/torque based learning for object manipulation.
\newblock {\em Robotics and Computer-Integrated Manufacturing}, 69:102111, 2021.

\bibitem{martinez2019concise}
Lourdes Mart{\'\i}nez-Villase{\~n}or and Hiram Ponce.
\newblock A concise review on sensor signal acquisition and transformation applied to human activity recognition and human--robot interaction.
\newblock {\em International Journal of Distributed Sensor Networks}, 15(6):1550147719853987, 2019.

\bibitem{cao2017realtime}
Zhe Cao, Tomas Simon, Shih-En Wei, and Yaser Sheikh.
\newblock Realtime multi-person 2d pose estimation using part affinity fields.
\newblock In {\em Proceedings of the IEEE conference on computer vision and pattern recognition}, pages 7291--7299, 2017.

\bibitem{cheng20203d}
Yu~Cheng, Bo~Yang, Bo~Wang, and Robby~T Tan.
\newblock 3d human pose estimation using spatio-temporal networks with explicit occlusion training.
\newblock In {\em Proceedings of the AAAI Conference on Artificial Intelligence}, volume~34, pages 10631--10638, 2020.

\bibitem{lee2018propagating}
Kyoungoh Lee, Inwoong Lee, and Sanghoon Lee.
\newblock Propagating lstm: 3d pose estimation based on joint interdependency.
\newblock In {\em Proceedings of the European conference on computer vision (ECCV)}, pages 119--135, 2018.

\bibitem{chen2019unsupervised}
Ching-Hang Chen, Ambrish Tyagi, Amit Agrawal, Dylan Drover, Rohith Mv, Stefan Stojanov, and James~M Rehg.
\newblock Unsupervised 3d pose estimation with geometric self-supervision.
\newblock In {\em Proceedings of the IEEE/CVF Conference on Computer Vision and Pattern Recognition}, pages 5714--5724, 2019.

\bibitem{Cao2021.OpenPose}
Z.~Cao, G.~Hidalgo, T.~Simon, S.~Wei, and Y.~Sheikh.
\newblock Openpose: Realtime multi-person 2d pose estimation using part affinity fields.
\newblock {\em IEEE Transactions on Pattern Analysis \& Machine Intelligence}, 43(01):172--186, jan 2021.

\bibitem{10000133}
Jan Docekal, Jakub Rozlivek, Jiri Matas, and Matej Hoffmann.
\newblock Human keypoint detection for close proximity human-robot interaction.
\newblock In {\em 2022 IEEE-RAS 21st International Conference on Humanoid Robots (Humanoids)}, pages 450--457, 2022.

\bibitem{fujii2018gaze}
Kenko Fujii, Gauthier Gras, Antonino Salerno, and Guang-Zhong Yang.
\newblock Gaze gesture based human robot interaction for laparoscopic surgery.
\newblock {\em Medical image analysis}, 44:196--214, 2018.

\bibitem{dermy2019multi}
Oriane Dermy, Fran{\c{c}}ois Charpillet, and Serena Ivaldi.
\newblock Multi-modal intention prediction with probabilistic movement primitives.
\newblock In {\em Human Friendly Robotics: 10th International Workshop}, pages 181--196. Springer, 2019.

\bibitem{Mazhar2018.HandGestures}
Osama Mazhar, Sofiane Ramdani, Benjamin Navarro, Robin Passama, and Andrea Cherubini.
\newblock Towards real-time physical human-robot interaction using skeleton information and hand gestures.
\newblock In {\em 2018 IEEE/RSJ International Conference on Intelligent Robots and Systems (IROS)}, pages 1--6, 2018.

\bibitem{mazhar2019real}
Osama Mazhar, Benjamin Navarro, Sofiane Ramdani, Robin Passama, and Andrea Cherubini.
\newblock A real-time human-robot interaction framework with robust background invariant hand gesture detection.
\newblock {\em Robotics and Computer-Integrated Manufacturing}, 60:34--48, 2019.

\bibitem{roda2021human}
Luis Roda-Sanchez, Teresa Olivares, Celia Garrido-Hidalgo, Jos{\'e}~Luis de~la Vara, and Antonio Fernandez-Caballero.
\newblock Human-robot interaction in industry 4.0 based on an internet of things real-time gesture control system.
\newblock {\em Integrated Computer-Aided Engineering}, 28(2):159--175, 2021.

\bibitem{Romano2018.CoDyCo}
Francesco Romano, Gabriele Nava, Morteza Azad, Jernej Čamernik, Stefano Dafarra, Oriane Dermy, Claudia Latella, Maria Lazzaroni, Ryan Lober, Marta Lorenzini, Daniele Pucci, Olivier Sigaud, Silvio Traversaro, Jan Babič, Serena Ivaldi, Michael Mistry, Vincent Padois, and Francesco Nori.
\newblock The codyco project achievements and beyond: Toward human aware whole-body controllers for physical human robot interaction.
\newblock {\em IEEE Robotics and Automation Letters}, 3(1):516--523, 2018.

\bibitem{9447230}
Lorenzo Vianello, Jean-Baptiste Mouret, Eloise Dalin, Alexis Aubry, and Serena Ivaldi.
\newblock Human posture prediction during physical human-robot interaction.
\newblock {\em IEEE Robotics and Automation Letters}, 6(3):6046--6053, 2021.

\bibitem{lanini2018human}
Jessica Lanini, Hamed Razavi, Julen Urain, and Auke Ijspeert.
\newblock Human intention detection as a multiclass classification problem: Application in physical human--robot interaction while walking.
\newblock {\em IEEE Robotics and Automation Letters}, 3(4):4171--4178, 2018.

\bibitem{liu2019intention}
Zhiguang Liu and Jianhong Hao.
\newblock Intention recognition in physical human-robot interaction based on radial basis function neural network.
\newblock {\em Journal of Robotics}, 2019, 2019.

\bibitem{8701608}
Guangzhu Peng, Chenguang Yang, Wei He, and C.~L.~Philip Chen.
\newblock Force sensorless admittance control with neural learning for robots with actuator saturation.
\newblock {\em IEEE Transactions on Industrial Electronics}, 67(4):3138--3148, 2020.

\bibitem{vianello2021human}
Lorenzo Vianello, Luigi Penco, Waldez Gomes, Yang You, Salvatore~Maria Anzalone, Pauline Maurice, Vincent Thomas, and Serena Ivaldi.
\newblock Human-humanoid interaction and cooperation: a review.
\newblock {\em Current Robotics Reports}, 2(4):441--454, 2021.

\bibitem{6386040}
Martin Lawitzky, José~Ramón Medina, Dongheui Lee, and Sandra Hirche.
\newblock Feedback motion planning and learning from demonstration in physical robotic assistance: differences and synergies.
\newblock In {\em 2012 IEEE/RSJ International Conference on Intelligent Robots and Systems}, pages 3646--3652, 2012.

\bibitem{li2021hybrid}
Shiqi Li, Ke~Han, Xiao Li, Shuai Zhang, Youjun Xiong, and Zheng Xie.
\newblock Hybrid trajectory replanning-based dynamic obstacle avoidance for physical human-robot interaction.
\newblock {\em Journal of Intelligent \& Robotic Systems}, 103:1--14, 2021.

\bibitem{8869047}
Marco Faroni, Manuel Beschi, and Nicola Pedrocchi.
\newblock An mpc framework for online motion planning in human-robot collaborative tasks.
\newblock In {\em 2019 24th IEEE International Conference on Emerging Technologies and Factory Automation (ETFA)}, pages 1555--1558, 2019.

\bibitem{moon2021design}
Ajung Moon, Maneezhay Hashmi, HF~Machiel Van~Der Loos, Elizabeth~A Croft, and Aude Billard.
\newblock Design of hesitation gestures for nonverbal human-robot negotiation of conflicts.
\newblock {\em ACM Transactions on Human-Robot Interaction (THRI)}, 10(3):1--25, 2021.

\bibitem{chernova2014robot}
Sonia Chernova and Andrea~L Thomaz.
\newblock Robot learning from human teachers.
\newblock {\em Synthesis lectures on artificial intelligence and machine learning}, 8(3):1--121, 2014.

\bibitem{6907265}
Heni Ben~Amor, Gerhard Neumann, Sanket Kamthe, Oliver Kroemer, and Jan Peters.
\newblock Interaction primitives for human-robot cooperation tasks.
\newblock In {\em 2014 IEEE International Conference on Robotics and Automation (ICRA)}, pages 2831--2837, 2014.

\bibitem{lai2022user}
Yujun Lai, Gavin Paul, Yunduan Cui, and Takamitsu Matsubara.
\newblock User intent estimation during robot learning using physical human robot interaction primitives.
\newblock {\em Autonomous Robots}, 46(2):421--436, 2022.

\bibitem{niekum2012learning}
Scott Niekum, Sarah Osentoski, George Konidaris, and Andrew~G Barto.
\newblock Learning and generalization of complex tasks from unstructured demonstrations.
\newblock In {\em 2012 IEEE/RSJ International Conference on Intelligent Robots and Systems}, pages 5239--5246. IEEE, 2012.

\bibitem{7451881}
Nadia Figueroa, Ana Lucia~Pais Ureche, and Aude Billard.
\newblock Learning complex sequential tasks from demonstration: A pizza dough rolling case study.
\newblock In {\em 2016 11th ACM/IEEE International Conference on Human-Robot Interaction (HRI)}, pages 611--612, 2016.

\bibitem{8607032}
Rui Huang, Hong Cheng, Jing Qiu, and Jianwei Zhang.
\newblock Learning physical human–robot interaction with coupled cooperative primitives for a lower exoskeleton.
\newblock {\em IEEE Transactions on Automation Science and Engineering}, 16(4):1566--1574, 2019.

\bibitem{losey2022physical}
Dylan~P Losey, Andrea Bajcsy, Marcia~K O’Malley, and Anca~D Dragan.
\newblock Physical interaction as communication: Learning robot objectives online from human corrections.
\newblock {\em The International Journal of Robotics Research}, 41(1):20--44, 2022.

\bibitem{losey2018review}
Dylan~P Losey, Craig~G McDonald, Edoardo Battaglia, and Marcia~K O'Malley.
\newblock A review of intent detection, arbitration, and communication aspects of shared control for physical human--robot interaction.
\newblock {\em Applied Mechanics Reviews}, 70(1), 2018.

\bibitem{9501975}
Mario Selvaggio, Marco Cognetti, Stefanos Nikolaidis, Serena Ivaldi, and Bruno Siciliano.
\newblock Autonomy in physical human-robot interaction: A brief survey.
\newblock {\em IEEE Robotics and Automation Letters}, 6(4):7989--7996, 2021.

\bibitem{khan2014compliance}
Said~G Khan, Guido Herrmann, Mubarak Al~Grafi, Tony Pipe, and Chris Melhuish.
\newblock Compliance control and human--robot interaction: Part 1—survey.
\newblock {\em International journal of humanoid robotics}, 11(03):1430001, 2014.

\bibitem{li2020control}
Hsieh-Yu Li, Audelia~G Dharmawan, Ishara Paranawithana, Liangjing Yang, and U-Xuan Tan.
\newblock A control scheme for physical human-robot interaction coupled with an environment of unknown stiffness.
\newblock {\em Journal of Intelligent \& Robotic Systems}, 100:165--182, 2020.

\bibitem{sharifi2014nonlinear}
Mojtaba Sharifi, Saeed Behzadipour, and Gholamreza Vossoughi.
\newblock Nonlinear model reference adaptive impedance control for human--robot interactions.
\newblock {\em Control Engineering Practice}, 32:9--27, 2014.

\bibitem{rhee2023hybrid}
Issac Rhee, Gitae Kang, Seung~Jae Moon, Yun~Seok Choi, and Hyouk~Ryeol Choi.
\newblock Hybrid impedance and admittance control of robot manipulator with unknown environment.
\newblock {\em Intelligent Service Robotics}, 16(1):49--60, 2023.

\bibitem{haninger2022model}
Kevin Haninger, Christian Hegeler, and Luka Peternel.
\newblock Model predictive control with gaussian processes for flexible multi-modal physical human robot interaction.
\newblock In {\em 2022 International Conference on Robotics and Automation (ICRA)}, pages 6948--6955. IEEE, 2022.

\bibitem{7934303}
Bryan Whitsell and Panagiotis Artemiadis.
\newblock Physical human–robot interaction (phri) in 6 dof with asymmetric cooperation.
\newblock {\em IEEE Access}, 5:10834--10845, 2017.

\bibitem{brahmi2019compliant}
Brahim Brahmi, Mohamed~Hamza Laraki, Maarouf Saad, Cristobal Ochoa-Luna, and Abdelkrim Brahmi.
\newblock Compliant adaptive control of human upper-limb exoskeleton robot with unknown dynamics based on a modified function approximation technique (mfat).
\newblock {\em Robotics and Autonomous Systems}, 117:92--102, 2019.

\bibitem{keemink2018admittance}
Arvid~QL Keemink, Herman van~der Kooij, and Arno~HA Stienen.
\newblock Admittance control for physical human--robot interaction.
\newblock {\em The International Journal of Robotics Research}, 37(11):1421--1444, 2018.

\bibitem{huang1992compliant}
H-P Huang and S-S Chen.
\newblock Compliant motion control of robots by using variable impedance.
\newblock {\em The International Journal of Advanced Manufacturing Technology}, 7(6):322--332, 1992.

\bibitem{9325872}
Mojtaba Sharifi, Amir Zakerimanesh, Javad~K. Mehr, Ali Torabi, Vivian~K. Mushahwar, and Mahdi Tavakoli.
\newblock Impedance variation and learning strategies in human–robot interaction.
\newblock {\em IEEE Transactions on Cybernetics}, 52(7):6462--6475, 2022.

\bibitem{lee2022real}
Kyeong~Ha Lee, Seung~Guk Baek, Hyuk~Jin Lee, Seung~Ho Lee, and Ja~Choon Koo.
\newblock Real-time adaptive impedance compensator using simultaneous perturbation stochastic approximation for enhanced physical human--robot interaction transparency.
\newblock {\em Robotics and Autonomous Systems}, 147:103916, 2022.

\bibitem{yu2020simplified}
Wen Yu and Adolfo Perrusqu{\'\i}a.
\newblock Simplified stable admittance control using end-effector orientations.
\newblock {\em International Journal of Social Robotics}, 12(5):1061--1073, 2020.

\bibitem{dong2020physical}
Jianwei Dong, Jianming Xu, Qiaoqian Zhou, and Songda Hu.
\newblock Physical human--robot interaction force control method based on adaptive variable impedance.
\newblock {\em Journal of the Franklin Institute}, 357(12):7864--7878, 2020.

\bibitem{8093992}
Francesco Romano, Gabriele Nava, Morteza Azad, Jernej Čamernik, Stefano Dafarra, Oriane Dermy, Claudia Latella, Maria Lazzaroni, Ryan Lober, Marta Lorenzini, Daniele Pucci, Olivier Sigaud, Silvio Traversaro, Jan Babič, Serena Ivaldi, Michael Mistry, Vincent Padois, and Francesco Nori.
\newblock The codyco project achievements and beyond: Toward human aware whole-body controllers for physical human robot interaction.
\newblock {\em IEEE Robotics and Automation Letters}, 3(1):516--523, 2018.

\bibitem{8463167}
Kazuya Otani, Karim Bouyarmane, and Serena Ivaldi.
\newblock Generating assistive humanoid motions for co-manipulation tasks with a multi-robot quadratic program controller.
\newblock In {\em 2018 IEEE International Conference on Robotics and Automation (ICRA)}, pages 3107--3113, 2018.

\bibitem{8624995}
Marie Charbonneau, Valerio Modugno, Francesco Nori, Giuseppe Oriolo, Daniele Pucci, and Serena Ivaldi.
\newblock Learning robust task priorities of qp-based whole-body torque-controllers.
\newblock In {\em 2018 IEEE-RAS 18th International Conference on Humanoid Robots (Humanoids)}, pages 1--9, 2018.

\bibitem{tassi2022adaptive}
Francesco Tassi, Elena De~Momi, and Arash Ajoudani.
\newblock An adaptive compliance hierarchical quadratic programming controller for ergonomic human--robot collaboration.
\newblock {\em Robotics and Computer-Integrated Manufacturing}, 78:102381, 2022.

\bibitem{tassi2023multi}
Francesco Tassi and Arash Ajoudani.
\newblock Multi-modal and adaptive control of human-robot interaction through hierarchical quadratic programming.
\newblock 2023.

\bibitem{hoffman2018whole}
Enrico~Mingo Hoffman, Brice Clement, Chengxu Zhou, Nikos~G Tsagarakis, Jean-Baptiste Mouret, and Serena Ivaldi.
\newblock Whole-body compliant control of icub: first results with opensot.
\newblock In {\em IEEE/RAS ICRA Workshop on Dynamic Legged Locomotion in Realistic Terrains}, 2018.

\bibitem{7803289}
Diego~Felipe Paez~Granados, Jun Kinugawa, Yasuhisa Hirata, and Kazuhiro Kosuge.
\newblock Guiding human motions in physical human-robot interaction through com motion control of a dance teaching robot.
\newblock In {\em 2016 IEEE-RAS 16th International Conference on Humanoid Robots (Humanoids)}, pages 279--285, 2016.

\bibitem{pugach2016touch}
Ganna Pugach, Artem Melnyk, Olga Tolochko, Alexandre Pitti, and Philippe Gaussier.
\newblock Touch-based admittance control of a robotic arm using neural learning of an artificial skin.
\newblock In {\em 2016 IEEE/RSJ International Conference on Intelligent Robots and Systems (IROS)}, pages 3374--3380. IEEE, 2016.

\bibitem{8884133}
Sven Cremer, Sumit~Kumar Das, Indika~B. Wijayasinghe, Dan~O. Popa, and Frank~L. Lewis.
\newblock Model-free online neuroadaptive controller with intent estimation for physical human–robot interaction.
\newblock {\em IEEE Transactions on Robotics}, 36(1):240--253, 2020.

\bibitem{app11125459}
Nguyen-Van Toan, Phan-Bui Khoi, and Soo-Yeong Yi.
\newblock A mlp-hedge-algebras admittance controller for physical human–robot interaction.
\newblock {\em Applied Sciences}, 11(12), 2021.

\bibitem{9548781}
Xinbo Yu, Bin Li, Wei He, Yanghe Feng, Long Cheng, and Carlos Silvestre.
\newblock Adaptive-constrained impedance control for human–robot co-transportation.
\newblock {\em IEEE Transactions on Cybernetics}, 52(12):13237--13249, 2022.

\bibitem{10173753}
Xinbo Yu, Sisi Liu, Shuang Zhang, Wei He, and Haifeng Huang.
\newblock Adaptive neural network force tracking control of flexible joint robot with an uncertain environment.
\newblock {\em IEEE Transactions on Industrial Electronics}, 71(6):5941--5949, 2024.

\bibitem{9666912}
Dong Wei, Lipeng Chen, Longfei Zhao, Hua Zhou, and Bidan Huang.
\newblock A vision-based measure of environmental effects on inferring human intention during human robot interaction.
\newblock {\em IEEE Sensors Journal}, 22(5):4246--4256, 2022.

\bibitem{roveda2020model}
Loris Roveda, Jeyhoon Maskani, Paolo Franceschi, Arash Abdi, Francesco Braghin, Lorenzo Molinari~Tosatti, and Nicola Pedrocchi.
\newblock Model-based reinforcement learning variable impedance control for human-robot collaboration.
\newblock {\em Journal of Intelligent \& Robotic Systems}, 100(2):417--433, 2020.

\bibitem{9786637}
Ruofan Wu, Minhan Li, Zhikai Yao, Wentao Liu, Jennie Si, and He~Huang.
\newblock Reinforcement learning impedance control of a robotic prosthesis to coordinate with human intact knee motion.
\newblock {\em IEEE Robotics and Automation Letters}, 7(3):7014--7020, 2022.

\bibitem{7353494}
Fotios Dimeas and Nikos Aspragathos.
\newblock Reinforcement learning of variable admittance control for human-robot co-manipulation.
\newblock In {\em 2015 IEEE/RSJ International Conference on Intelligent Robots and Systems (IROS)}, pages 1011--1016, 2015.

\bibitem{lillicrap2015continuous}
Timothy~P Lillicrap, Jonathan~J Hunt, Alexander Pritzel, Nicolas Heess, Tom Erez, Yuval Tassa, David Silver, and Daan Wierstra.
\newblock Continuous control with deep reinforcement learning.
\newblock {\em arXiv preprint arXiv:1509.02971}, 2015.

\bibitem{liu2021deep}
Quan Liu, Zhihao Liu, Bo~Xiong, Wenjun Xu, and Yang Liu.
\newblock Deep reinforcement learning-based safe interaction for industrial human-robot collaboration using intrinsic reward function.
\newblock {\em Advanced Engineering Informatics}, 49:101360, 2021.

\bibitem{9849514}
Jong~In Han, Jeong-Hoon Lee, Ho~Seon Choi, Jung-Hoon Kim, and Jongeun Choi.
\newblock Policy design for an ankle-foot orthosis using simulated physical human–robot interaction via deep reinforcement learning.
\newblock {\em IEEE Transactions on Neural Systems and Rehabilitation Engineering}, 30:2186--2197, 2022.

\bibitem{9531394}
Mojtaba Sharifi, Vahid Azimi, Vivian~K. Mushahwar, and Mahdi Tavakoli.
\newblock Impedance learning-based adaptive control for human–robot interaction.
\newblock {\em IEEE Transactions on Control Systems Technology}, 30(4):1345--1358, 2022.

\bibitem{6224877}
Klas Kronander and Aude Billard.
\newblock Online learning of varying stiffness through physical human-robot interaction.
\newblock In {\em 2012 IEEE International Conference on Robotics and Automation}, pages 1842--1849, 2012.

\bibitem{kronander2013learning}
Klas Kronander and Aude Billard.
\newblock Learning compliant manipulation through kinesthetic and tactile human-robot interaction.
\newblock {\em IEEE transactions on haptics}, 7(3):367--380, 2013.

\bibitem{9842322}
Wentao Liu, Ruofan Wu, Jennie Si, and He~Huang.
\newblock A new robotic knee impedance control parameter optimization method facilitated by inverse reinforcement learning.
\newblock {\em IEEE Robotics and Automation Letters}, 7(4):10882--10889, 2022.

\bibitem{thompson2020computational}
Neil~C Thompson, Kristjan Greenewald, Keeheon Lee, and Gabriel~F Manso.
\newblock The computational limits of deep learning.
\newblock {\em arXiv preprint arXiv:2007.05558}, 2020.

\bibitem{baskakov2021computational}
Dmitry Baskakov and Dmitry Arseniev.
\newblock On the computational complexity of deep learning algorithms.
\newblock In {\em Proceedings of International Scientific Conference on Telecommunications, Computing and Control: TELECCON 2019}, pages 343--356. Springer, 2021.

\bibitem{dally2021evolution}
William~J Dally, Stephen~W Keckler, and David~B Kirk.
\newblock Evolution of the graphics processing unit (gpu).
\newblock {\em IEEE Micro}, 41(6):42--51, 2021.

\bibitem{9774600}
Rubens~Luiz Rech and Paolo Rech.
\newblock Reliability of google's tensor processing units for embedded applications.
\newblock In {\em 2022 Design, Automation \& Test in Europe Conference \& Exhibition (DATE)}, pages 376--381, 2022.

\bibitem{1708640}
P.H.W. Leong and K.H. Tsoi.
\newblock Field programmable gate array technology for robotics applications.
\newblock In {\em 2005 IEEE International Conference on Robotics and Biomimetics - ROBIO}, pages 295--298, 2005.

\bibitem{9350173}
Brian Plancher, Sabrina~M. Neuman, Thomas Bourgeat, Scott Kuindersma, Srinivas Devadas, and Vijay~Janapa Reddi.
\newblock Accelerating robot dynamics gradients on a cpu, gpu, and fpga.
\newblock {\em IEEE Robotics and Automation Letters}, 6(2):2335--2342, 2021.

\bibitem{tu2019power}
Yuexuan Tu, Saad Sadiq, Yudong Tao, Mei-Ling Shyu, and Shu-Ching Chen.
\newblock A power efficient neural network implementation on heterogeneous fpga and gpu devices.
\newblock In {\em 2019 IEEE 20th international conference on information reuse and integration for data science (IRI)}, pages 193--199. IEEE, 2019.

\bibitem{kortenkamp2016robotic}
David Kortenkamp, Reid Simmons, and Davide Brugali.
\newblock Robotic systems architectures and programming.
\newblock {\em Springer handbook of robotics}, pages 283--306, 2016.

\bibitem{ingrand2017deliberation}
F{\'e}lix Ingrand and Malik Ghallab.
\newblock Deliberation for autonomous robots: A survey.
\newblock {\em Artificial Intelligence}, 247:10--44, 2017.

\bibitem{4048968}
A.~Meystel.
\newblock Planning in a hierarchical nested controller for autonomous robots.
\newblock In {\em 1986 25th IEEE Conference on Decision and Control}, pages 1237--1249, 1986.

\bibitem{albus1995nist}
James~S Albus et~al.
\newblock The nist real-time control system (rcs): An application survey.
\newblock In {\em Proc. of the AAAI 1995 Spring Symposium Series, Stanford University, Menlo Park, CA}, 1995.

\bibitem{arkin1998behavior}
Ronald~C Arkin.
\newblock {\em Behavior-based robotics}.
\newblock MIT press, 1998.

\bibitem{qureshi2004cognitive}
Faisal Qureshi, Demetri Terzopoulos, and Ross Gillett.
\newblock The cognitive controller: a hybrid, deliberative/reactive control architecture for autonomous robots.
\newblock In {\em International Conference on Industrial, Engineering and Other Applications of Applied Intelligent Systems}, pages 1102--1111. Springer, 2004.

\bibitem{1087032}
R.~Brooks.
\newblock A robust layered control system for a mobile robot.
\newblock {\em IEEE Journal on Robotics and Automation}, 2(1):14--23, 1986.

\bibitem{horswill1993polly}
Ian Horswill.
\newblock Polly: A vision-based artificial agent.

\bibitem{mataric2018integration}
Maja~J Matari{\'c}.
\newblock Integration of representation into goal-driven behavior-based robots.
\newblock In {\em The artificial life route to artificial intelligence}, pages 165--186. Routledge, 2018.

\bibitem{toal1996subsumption}
Daniel Toal, Colin Flanagan, Caimin Jones, and Bob Strunz.
\newblock Subsumption architecture for the control of robots.
\newblock {\em IMC-13, Limerick}, 1996.

\bibitem{amoretti2010architectural}
Michele Amoretti and Monica Reggiani.
\newblock Architectural paradigms for robotics applications.
\newblock {\em Advanced Engineering Informatics}, 24(1):4--13, 2010.

\bibitem{chavan2015review}
PU~Chavan, M~Murugan, and PP~Chavan.
\newblock A review on software architecture styles with layered robotic software architecture.
\newblock In {\em 2015 International Conference on Computing Communication Control and Automation}, pages 827--831. IEEE, 2015.

\bibitem{quigley2009ros}
Morgan Quigley, Ken Conley, Brian Gerkey, Josh Faust, Tully Foote, Jeremy Leibs, Rob Wheeler, Andrew~Y Ng, et~al.
\newblock Ros: an open-source robot operating system.
\newblock In {\em ICRA workshop on open source software}, volume~3, page~5. Kobe, Japan, 2009.

\bibitem{martinetti2021redefining}
Alberto Martinetti, Peter~K Chemweno, Kostas Nizamis, and Eduard Fosch-Villaronga.
\newblock Redefining safety in light of human-robot interaction: A critical review of current standards and regulations.
\newblock {\em Frontiers in chemical engineering}, 3:32, 2021.

\bibitem{riek2014codeOfEthics}
Laurel Riek and Don Howard.
\newblock A code of ethics for the human-robot interaction profession.
\newblock {\em Proceedings of we robot}, 2014.

\bibitem{9162045}
Yusuf Aydin, Ozan Tokatli, Volkan Patoglu, and Cagatay Basdogan.
\newblock A computational multicriteria optimization approach to controller design for physical human-robot interaction.
\newblock {\em IEEE Transactions on Robotics}, 36(6):1791--1804, 2020.

\bibitem{setchi2020explainable}
Rossitza Setchi, Maryam~Banitalebi Dehkordi, and Juwairiya~Siraj Khan.
\newblock Explainable robotics in human-robot interactions.
\newblock {\em Procedia Computer Science}, 176:3057--3066, 2020.

\bibitem{lera2017cybersecurity}
Francisco J~Rodr{\'\i}guez Lera, Camino~Fern{\'a}ndez Llamas, {\'A}ngel~Manuel Guerrero, and Vicente~Matell{\'a}n Olivera.
\newblock Cybersecurity of robotics and autonomous systems: Privacy and safety.
\newblock {\em Robotics-legal, ethical and socioeconomic impacts}, 2017.

\bibitem{Hamacher2016TrustpHRI}
Adriana Hamacher, Nadia Bianchi-Berthouze, Anthony~G. Pipe, and Kerstin Eder.
\newblock Believing in bert: Using expressive communication to enhance trust and counteract operational error in physical human-robot interaction.
\newblock In {\em 2016 25th IEEE International Symposium on Robot and Human Interactive Communication (RO-MAN)}, pages 493--500, 2016.

\bibitem{9698842}
Yue Hu, Naoko Abe, Mehdi Benallegue, Natsuki Yamanobe, Gentiane Venture, and Eiichi Yoshida.
\newblock Toward active physical human–robot interaction: Quantifying the human state during interactions.
\newblock {\em IEEE Transactions on Human-Machine Systems}, 52(3):367--378, 2022.

\bibitem{fronemann2022EthicsUX}
Nora Fronemann, Kathrin Pollmann, and Wulf Loh.
\newblock Should my robot know what's best for me? human--robot interaction between user experience and ethical design.
\newblock {\em AI \& SOCIETY}, 37(2):517--533, 2022.

\bibitem{howard2018bias}
Ayanna Howard and Jason Borenstein.
\newblock The ugly truth about ourselves and our robot creations: the problem of bias and social inequity.
\newblock {\em Science and engineering ethics}, 24:1521--1536, 2018.

\bibitem{williams2023eye}
Tom Williams.
\newblock The eye of the robot beholder: Ethical risks of representation, recognition, and reasoning over identity characteristics in human-robot interaction.
\newblock In {\em Companion of the 2023 ACM/IEEE International Conference on Human-Robot Interaction}, HRI '23, page 1–10, New York, NY, USA, 2023. Association for Computing Machinery.

\bibitem{meissner2020friend}
Antonia Meissner, Angelika Tr{\"u}bswetter, Antonia~S Conti-Kufner, and Jonas Schmidtler.
\newblock Friend or foe? understanding assembly workers’ acceptance of human-robot collaboration.
\newblock {\em ACM Transactions on Human-Robot Interaction (THRI)}, 10(1):1--30, 2020.

\bibitem{van2022ethical}
Aimee van Wynsberghe, Madelaine Ley, and Sabine Roeser.
\newblock Ethical aspects of human--robot collaboration in industrial work settings.
\newblock {\em The 21st Century Industrial Robot: When Tools Become Collaborators}, pages 255--266, 2022.

\bibitem{etemad2022ethical}
Reza Etemad-Sajadi, Antonin Soussan, and Th{\'e}o Sch{\"o}pfer.
\newblock How ethical issues raised by human--robot interaction can impact the intention to use the robot?
\newblock {\em International journal of social robotics}, 14(4):1103--1115, 2022.

\bibitem{moon2021ethics}
AJung Moon, Shalaleh Rismani, and HF~Machiel Van~der Loos.
\newblock Ethics of corporeal, co-present robots as agents of influence: a review.
\newblock {\em Current Robotics Reports}, 2:223--229, 2021.

\bibitem{fronemann2022should}
Nora Fronemann, Kathrin Pollmann, and Wulf Loh.
\newblock Should my robot know what's best for me? human--robot interaction between user experience and ethical design.
\newblock {\em AI \& SOCIETY}, 37(2):517--533, 2022.

\end{thebibliography}
\end{sloppypar}
\end{document}